\documentclass{article}

\usepackage[main, final]{neurips_2026}

\usepackage[utf8]{inputenc} 
\usepackage[T1]{fontenc}    
\usepackage{hyperref}       
\usepackage{url}            
\usepackage{booktabs}       
\usepackage{amsfonts}       
\usepackage{nicefrac}       
\usepackage{microtype}      
\usepackage{xcolor}         
\usepackage{graphicx}
\usepackage{algorithm}
\usepackage{algpseudocode}
\usepackage{amsmath}
\usepackage{multirow}
\usepackage{subcaption}

\usepackage{xcolor}
\usepackage[most]{tcolorbox}
\usepackage{enumitem}
\title{EpiStream: Utility-Aware Temporal Abstraction for Dense-Action Streams}

\author{%
  Tan Kai Ze$^{1,2}$%
  \thanks{Equal contribution.}\kern0.4em
  \quad
  Ran Zhang$^{2}$\footnotemark[1]\kern0.4em
  \thanks{Corresponding author.}\quad
  Feng Yichao$^{3}$\\
  \textbf{Matthew Heng Yu Jie}$^{3}$\quad
  \textbf{Dong Fang}$^{2}$\\[0.5em]
  $^{1}$National University of Singapore\\
  $^{2}$LightSpeed Studios, Tencent\\
  $^{3}$Nanyang Technological University
}

\begin{document}

\maketitle

\begingroup
\renewcommand{\thefootnote}{}
\footnotetext[0]{This work was conducted during internship at LightSpeed Studios, Tencent.}
\endgroup


\begin{abstract}
Video understanding is commonly built on action-triggered responses or scene-level aggregation, implicitly assuming clear temporal boundaries. This assumption breaks in dense-action streams such as gameplay and sports, where overlapping actions induce continuous latent dynamics without explicit transitions. The challenge is further amplified in streaming settings, where models operate under strict causal constraints and must progressively construct episodic memory for downstream reasoning. Existing segmentation-based strategies therefore often produce fragmented or semantically mixed episodes, leading to degraded understanding.
We present \textbf{EpiStream}, an online framework for utility-aware semantic episode formation in dense-action streams. Instead of detecting event boundaries, EpiStream learns when an evolving video prefix should be committed into an episode unit that maximizes downstream utility under limited memory budgets. The framework consists of three steps: (i) utility function design for semantic coherence, inter-episode distinctiveness, and downstream objectives; (ii) utility-optimal commit advantage construction from teaching signals; and (iii) peak-aware advantage learning for training a causal online commitment policy from prefix-only observations. Extensive experiments show that EpiStream substantially improves temporal semantic coherence and consistently benefits challenging downstream tasks, including real-time advice and player intention prediction. Our code is available at
\url{https://github.com/BryanTan0619/EpiStream}.
\end{abstract}

\section{Introduction}

Video understanding has been extensively studied through tasks such as action recognition, temporal segmentation, video captioning, and question answering~\cite{maaz2023videochatgpt,lin2023videollava,li2024mvbench,fu2024videomme,mangalam2023egoschema,wu2024longvideobench,zhou2024mlvu, heilbron2015activitynet, krishna2017densecaptioning}. Recent advances have further extended these problems to streaming scenarios, enabling real-time applications including online event detection, interactive commentary, and decision support~\cite{chen2024videollmonline,qian2024videostreaming,liu2024streamchat,chen2025livecc,wang2025streambridge,qian2025dispider}. Unlike offline video understanding, where models can access the entire sequence, streaming systems must operate causally over unbounded visual streams without access to future observations or the ability to revise past decisions. This challenge is particularly important in dense-action streams (DAS), such as competitive gameplay, sports broadcasts, and human activity streams, where overlapping actions induce continuous latent dynamics without explicit transitions~\cite{yeung2018everymoment,sardari2025refdense}. In such environments, the system must not only recognize ongoing events, but also determine when an evolving observation prefix forms a coherent episode suitable for downstream reasoning.

Despite its practical importance, DAS understanding remains underexplored and presents several fundamental challenges. \textbf{First}, dense-action streams often lack clear temporal boundaries, making trigger-based or scene-transition-based segmentation unreliable. \textbf{Second}, streaming systems must make irreversible commit decisions under causal constraints. Committing too early may produce incomplete semantic context, while committing too late may dilute critical information and reduce responsiveness. \textbf{Third}, effective temporal abstraction in DAS is inherently utility-dependent. A good episode should not only be semantically coherent and temporally distinctive, but also useful for downstream tasks such as player intention prediction or real-time advice. Existing methods typically optimize boundary accuracy or event detection performance, but do not explicitly model utility-aware episode formation under limited memory budgets.

\begin{figure}
  \centering
  \includegraphics[width=0.95\linewidth]{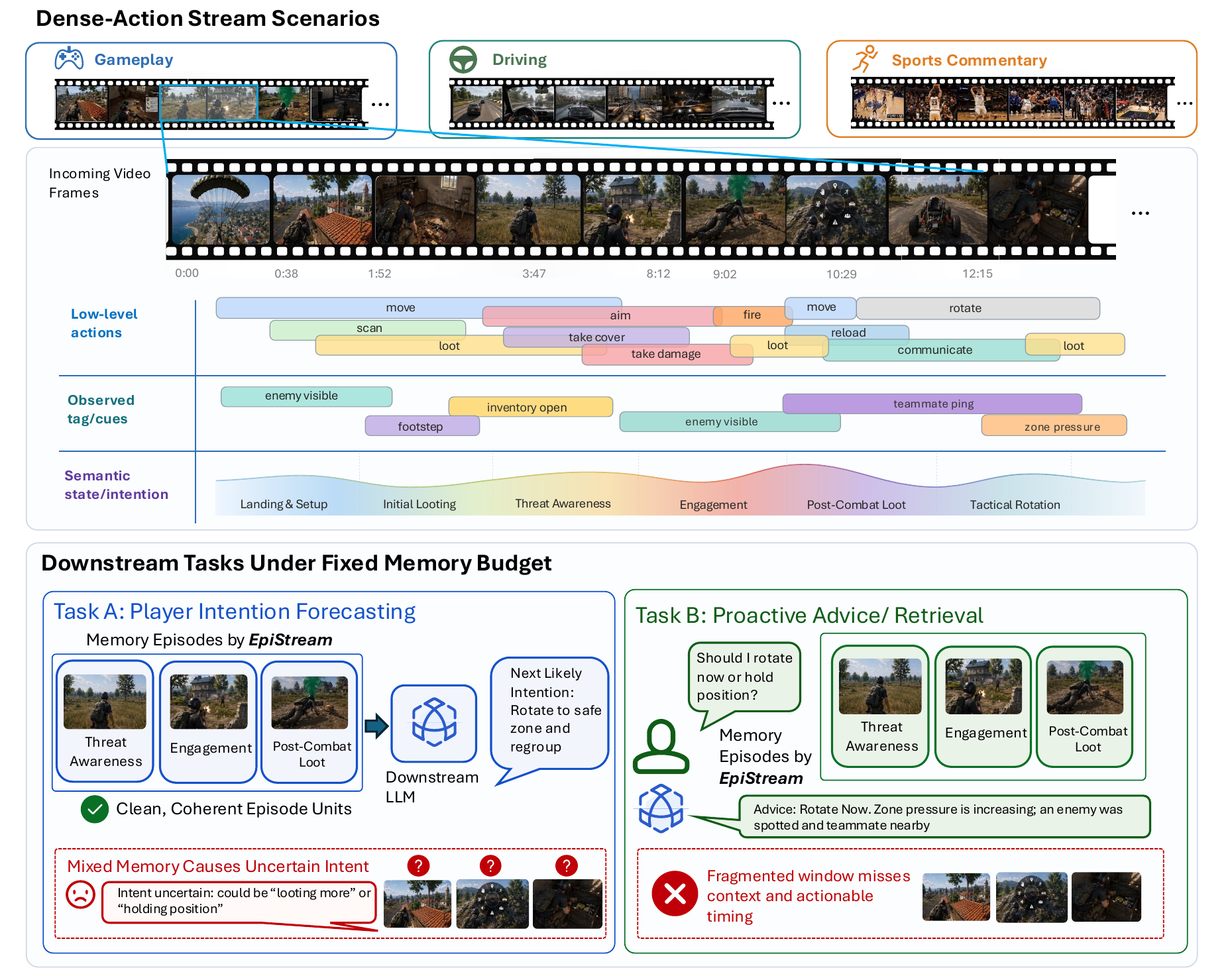}
  \caption{Motivation of EpiStream. Dense-action streams lack clear boundaries due to overlapping actions and evolving latent intentions. EpiStream forms coherent, utility-aware episodic units for downstream reasoning under a fixed memory budget.}
  \label{fig:figure_1}
\end{figure}

To address these challenges, we propose \textbf{EpiStream}, an online framework for utility-aware semantic episode formation in dense-action streams. Rather than treating streaming understanding as a boundary detection problem, EpiStream learns when an evolving video prefix should be committed into an episode unit that maximizes downstream utility. The framework consists of three components. First, we design a utility function that jointly captures semantic coherence, inter-episode distinctiveness, and downstream objectives. Second, using long-form gameplay streams with synchronized telemetry, we construct utility-optimal commit advantages as dense teaching signals for supervised learning. Third, we introduce a peak-aware advantage learning strategy that trains a lightweight causal commitment policy from prefix-only observations.

We evaluate EpiStream on long-form FPS gameplay streams with synchronized telemetry supervision. Experiments show that EpiStream produces substantially better temporal semantic coherence than heuristic and boundary-based baselines, while consistently improving downstream performance on player intention prediction and real-time advice. These results suggest that learning when to commit semantic episodes is a fundamental capability for dense-action streaming understanding.

\section{Related Work}

\paragraph{Streaming Video Understanding.}
Recent work has extended video understanding from offline, clip-based processing to online and streaming settings, where models operate over continuously arriving visual input under causal and real-time constraints. Representative efforts introduce temporally aligned training, stream-aware memory, continuous state tracking, and proactive interaction for sequential video understanding~\cite{chen2024videollmonline,qian2024videostreaming,liu2024streamchat,chen2025livecc,wang2025streambridge,xu2025streamingvlm}. These advances establish streaming video understanding as a distinct paradigm of continuous perception and interaction. However, they mainly focus on how to process or respond to a stream, leaving open how streaming observations should be temporally abstracted into semantically coherent episodes for downstream reasoning in dense-action settings.

\paragraph{Online Segmentation.}
A related line of work studies temporal decision making in streaming videos, including online segmentation, boundary detection, and query-conditioned response timing. Existing methods typically optimize objectives such as boundary correctness, event coherence, or response readiness, and have made important progress on organizing streams or triggering timely outputs~\cite{zhong2024onlinetas,jung2025onlinegebd,lu2026vista,guo2026eventvstream,azad2026streamready,qian2025dispider,zheng2026emgarde}. However, in dense-action streams, visually salient transitions and response-ready moments do not necessarily coincide with useful memory boundaries, leaving unresolved when a current prefix should be committed to maximize downstream utility.

\paragraph{Long-Video Memory, Retrieval, and Compression.}
Long-video understanding has also motivated extensive research on memory organization, retrieval, and compression, including persistent memory structures, learnable memory modules, and retrieval-based compression~\cite{zeng2025streamforest,yuan2025memvid,diko2025rewind,wang2025videoem,zuo2025videolucy,yin2025videoarm,azad2025hierarq}. These methods improve long-horizon reasoning by addressing what historical information to store, compress, or retrieve. However, they generally assume that an appropriate temporal unit is already available, while our setting requires deciding when an evolving prefix has become semantically coherent enough to be committed as a compact memory abstraction under causal and limited-budget constraints.

\begin{figure}
  \centering
  \includegraphics[width=0.95\linewidth]{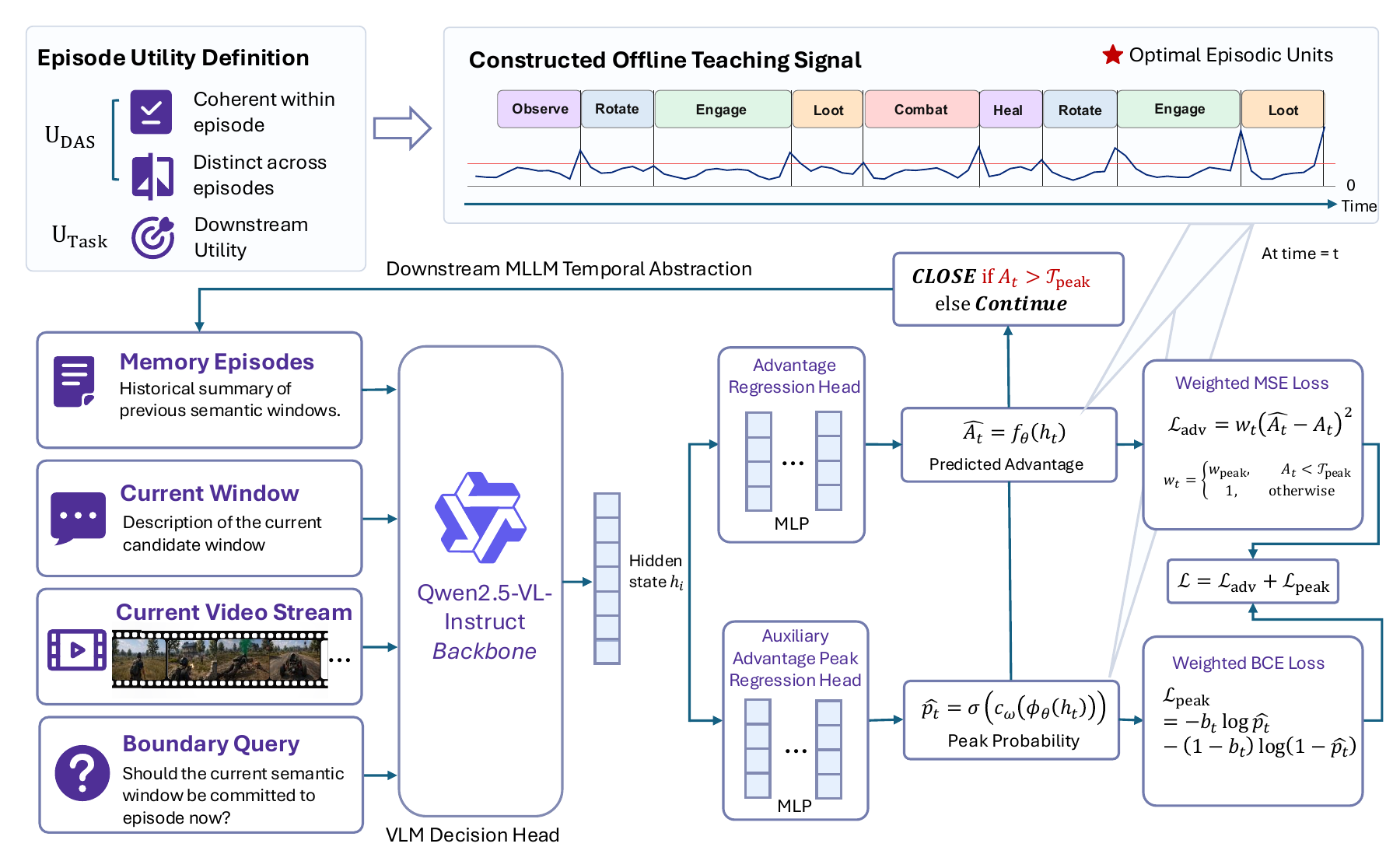}
  \caption{ Overview of \textbf{EpiStream}. Offline telemetry defines episode utility and close-versus-continue advantage targets. Online, a VLM encodes causal inputs into a decision state \(h_t\), from which coupled predictors estimate commit advantage \(\hat{A}_t\) and peak probability \(\hat{p}_t\). The policy is trained with a joint advantage-peak objective and commits an episode when \(\hat{A}_t\) exceeds the stopping threshold.
}
\label{fig:figure_2}
\end{figure}

\section{Methodology}
\label{sec:methodology}

In this section, we introduce \textbf{EpiStream}, an online framework for utility-aware semantic episode formation in dense-action streams. As shown in Fig.~\ref{fig:figure_2}, EpiStream learns when an evolving video prefix should be closed into an episode unit for downstream reasoning. The framework contains three components: (1) \emph{Utility Function Design}, which defines what makes an episode internally coherent, externally distinctive, and downstream-informative; (2) \emph{Offline Utility-Optimal Episode Construction}, which converts privileged telemetry data into teaching targets; and (3) \emph{Online Commitment Policy Learning}, which trains a online policy with coupled advantage and peak predictors to form structured episodic memory from prefix-only observations.

\subsection{Problem Formulation}
\label{sec:commit_formulation}

\paragraph{DAS Setting.}
Let \(X=\{x_1,\ldots,x_T\}\) denote a dense-action stream (DAS), where multiple actions may overlap and the underlying semantic state evolves continuously without explicit boundaries. We instantiate this setting on competitive gameplay streams, a representative DAS domain with synchronized telemetry for privileged supervision. At time \(t\), the model observes only the prefix \(X_{\le t}\), maintains an open episode candidate \(S_t=X_{\tau_t:t}\), and stores previously committed episodic memories \(M_t=\{m_1,\ldots,m_{k-1}\}\). The online policy chooses
\begin{equation}
a_t\in\{\textsc{Continue},\textsc{Commit}\}.
\end{equation}
If \(a_t=\textsc{Continue}\), the current candidate remains open; otherwise, it is committed into an episodic memory unit \(m_k=G(S_t)\), where \(G(\cdot)\) denotes the episodic memory constructor. The decision process is causal and irreversible: future observations and privileged signals are unavailable during inference, and committed episodes cannot be revised. The objective is therefore not boundary detection, but determining when the current prefix forms a semantically coherent and downstream-useful episode under limited memory budgets.





\paragraph{Episodic Memory Abstraction.}
EpiStream instantiates temporal abstraction as online episodic memory construction. Each committed episode is represented as
\begin{equation}
m_i=\{t_i^s,t_i^e,c_i,q_i,e_i,r_i\},
\end{equation}
where \(t_i^s,t_i^e\) denote the temporal span, \(c_i\) the episode phase, \(q_i\) a compact summary, \(e_i\) supporting evidence, and \(r_i\) an optional utility score. The episodic memory bank supports downstream reasoning and is recursively fed back into the online policy state. Note in all experiments, the episodic memory constructor \(G(\cdot)\), video budget, memory budget, and downstream model are fixed; only the episode formation policy changes.


\subsection{Offline Utility-Optimal Episode Construction}
\label{sec:epistream_framework}

EpiStream evaluates whether a committed episode forms a coherent and useful temporal abstraction for downstream reasoning. We decompose episode utility into a task-agnostic DAS utility and a task-oriented utility:
\begin{equation}
U(S_i)=
\lambda_{\mathrm{DAS}}U_{\mathrm{DAS}}(S_i)
+
\lambda_{\mathrm{task}}U_{\mathrm{task}}(S_i).
\end{equation}

The DAS utility captures two complementary aspects of semantic episode formation:
\begin{equation}
U_{\mathrm{DAS}}(S_i)=
\lambda_{\mathrm{sem}}U_{\mathrm{sem}}(S_i)
+
\lambda_{\mathrm{distinct}}U_{\mathrm{distinct}}(S_{i-1},S_i),
\end{equation}
where \(U_{\mathrm{sem}}\) measures semantic coherence within an episode and \(U_{\mathrm{distinct}}\) measures semantic distinctiveness across neighboring episodes. The task-oriented term \(U_{\mathrm{task}}\) adjusts the preferred temporal granularity for downstream objectives. Detailed utility terms are defined in Appendix~\ref{app:semantic_score_metrics}.

Given the utility definition, we construct offline utility-optimal episode references as teaching signals for online policy learning. The construction proceeds in three stages. First, we identify high-recall candidate commit points from synchronized telemetry signals by detecting timestamps associated with scene transitions, contextual shifts, and task-relevant changes in the underlying gameplay dynamics (examples are provided in Appendix~\ref{app:offline_utility_construction}). These candidates define plausible episode closure times in the stream. Second, for each candidate time \(t\), we compare the utility of committing the current episode against the utility of delaying the decision for future evidence accumulation within the same local reference window. This yields a close-versus-continue advantage:
\begin{equation}
A_t
=
U_{\mathrm{close}}(t)-U_{\mathrm{continue}}(t),
\end{equation}
where \(U_{\mathrm{close}}(t)=U(X_{\tau:t})\) evaluates the utility of committing at time \(t\), and
\begin{equation}
U_{\mathrm{continue}}(t)
=
\max_{c\in\mathcal{C}_t,\ c>t}
U_{\mathrm{close}}(c)
\end{equation}
estimates the best achievable utility by postponing the commit decision to a future candidate \(c\) within the same local window.

Finally, we convert these utility comparisons into dense advantage targets for policy training. Positive advantage indicates that the current prefix already forms a utility-optimal semantic episode, whereas negative advantage suggests that additional observations remain beneficial. Importantly, the resulting supervision is not based on human-annotated temporal boundaries, but on utility-derived reference decisions for episode formation.

\subsection{Online Policy Learning}

\begin{algorithm}[t]
\small
\caption{EpiStream Online Episode Formation}
\label{alg:epistream_online}
\begin{algorithmic}[1]
\Require Stream $X=\{x_t\}_{t=1}^{T}$; policy $f_\theta$; constructor $G$; thresholds $\gamma,d_{\min},d_{\max}$; budget $K$.
\Ensure Episodic memory bank $M$.

\State $M\gets\emptyset$, $\tau\gets 1$
\Comment{initialize memory and episode start}
\For{$t=1,\ldots,T$}
    \State $S_t\gets X_{\tau:t}$, $d_t\gets t-\tau+1$
    \Comment{current open episode candidate and duration}
    
    \State $h_t\gets \mathrm{BuildState}(M,S_t,d_t,x_{t-k+1:t})$
    \Comment{memory, candidate, duration, and recent $k$ frames}
    
    \State $\hat{A}_t\gets f_\theta(h_t)$
    \Comment{predict close-versus-continue advantage}
    
    \If{$(\hat{A}_t\ge\gamma \ \mathbf{and}\ d_t\ge d_{\min}) \ \mathbf{or}\ d_t\ge d_{\max}$}
        \State $m\gets G(S_t)$
        \Comment{construct episodic memory unit}
        
        \State $M\gets \mathrm{UpdateBudget}(M,m,K)$
        \Comment{update memory bank under budget}
        
        \State $\tau\gets t+1$
        \Comment{start a new open episode candidate}
    \Else
        \State \textbf{continue}
        \Comment{keep accumulating evidence}
    \EndIf
\EndFor
\State \Return $M$
\end{algorithmic}
\end{algorithm}

\subsubsection{Joint Advantage and Peak Modeling}
\label{sec:policy_learning}

The utility construction above turns offline privileged signals into close-versus-continue advantage targets, while the online policy must approximate them from causal observations only. This yields an optimal stopping problem: at each timestep, the model decides whether to close the current episode candidate or keep it open for more evidence. Although reinforcement learning is a natural formulation, it is inefficient here because useful closure actions are sparse, rewards are delayed, and credit assignment over long dense-action streams is unstable. We therefore train from utility-derived advantage supervision. Yet directly regressing \(A_t\) is not sufficient: the advantage landscape is highly imbalanced, with most timesteps lying in low-advantage \textsc{Continue} regions and utility-optimal closures appearing as sparse peaks. Pure regression can reduce global error while underestimating these decision-critical peaks. We view this problem by considering a training distribution decomposed into low-advantage and peak regions:

\begin{equation}
\mathcal{D}
=
(1-\pi)\mathcal{D}_{\mathrm{low}}
+
\pi\mathcal{D}_{\mathrm{peak}},
\qquad
\pi\ll 1.
\end{equation}

The population minimizer of mean squared error satisfies

\begin{equation}
f^\star(h)
=
\mathbb{E}[A_t\mid h_t=h]
=
(1-\pi_h)\mathbb{E}[A_t\mid h,\mathrm{low}]
+
\pi_h\mathbb{E}[A_t\mid h,\mathrm{peak}],
\end{equation}

where \(\pi_h=\mathbb{P}(\mathrm{peak}\mid h)\). When high-advantage samples are rare or not perfectly separable, \(f^\star(h)\) is pulled toward the dominant low-advantage region. Thus, a model can achieve low global regression error while underestimating the sparse peaks that determine correct episode-closure decisions.

To address this, EpiStream jointly models two complementary views of the same utility signal: the continuous commit advantage and the high-advantage closure region. At time \(t\), the policy receives a causal decision state \(h_t\), containing recent frames, the current open episode candidate, the existing episodic memory bank, and the current duration. It uses a shared representation \(\phi_\theta(h_t)\) with two coupled predictors:

\begin{equation}
\hat{A}_t
=
r_\psi(\phi_\theta(h_t)),
\qquad
\hat{p}_t
=
\sigma(c_\omega(\phi_\theta(h_t))).
\end{equation}

Here \(\hat{A}_t\) estimates the close-versus-continue advantage, while \(\hat{p}_t\) estimates whether the current timestep lies in a high-advantage episode-closure region. We define the binary peak target as
\(
b_t=\mathbb{I}(A_t\ge \eta),
\)
where \(\eta\) is a percentile threshold of the empirical advantage distribution. This peak target does not introduce a separate supervision source; it is derived from the same commit advantage \(A_t\) and provides a region-level view of where stopping decisions should occur.

We optimize a joint advantage-peak objective:

\begin{equation}
\begin{aligned}
\mathcal{L}
&=
\underbrace{
\mathbb{E}_t\!\left[
w_t(\hat{A}_t-A_t)^2
\right]
}_{\mathcal{L}_{\mathrm{adv}}}
+
\lambda_{\mathrm{peak}}
\underbrace{
\mathbb{E}_t\!\left[
-\alpha b_t\log \hat{p}_t
-(1-b_t)\log(1-\hat{p}_t)
\right]
}_{\mathcal{L}_{\mathrm{peak}}},\\
w_t
&=
\begin{cases}
w_{\mathrm{peak}}, & A_t\ge \eta_{\mathrm{peak}},\\
1, & \text{otherwise},
\end{cases}
\qquad
\alpha=\frac{N_{\mathrm{neg}}}{N_{\mathrm{pos}}}.
\end{aligned}
\end{equation}

The advantage term \(\mathcal{L}_{\mathrm{adv}}\) learns the magnitude of the close-versus-continue utility, with \(w_t\) upweighting rare high-advantage samples to counteract the dominance of \textsc{Continue} regions. The peak term \(\mathcal{L}_{\mathrm{peak}}\) performs class-balanced closure-region classification, encouraging the shared representation to separate sparse high-advantage closure regions from the low-advantage background. This objective can be viewed as regularized advantage regression, but our primary interpretation is joint modeling of utility magnitude and decision-critical closure regions. A probabilistic interpretation is provided in Appendix~\ref{app:offline_utility_construction}.

\subsubsection{Online Inference}
\label{sec:online_inference}

At inference time, EpiStream operates strictly online: telemetry, future frames, and offline reference targets are unavailable. The policy uses the predicted advantage \(\hat{A}_t\) as the stopping score. It closes the current episode candidate when the predicted advantage exceeds a threshold and the candidate is sufficiently long; closure is forced once the candidate reaches the maximum duration:
\begin{equation}
a_t =
\begin{cases}
\textsc{Commit}, & \hat{A}_t \ge \gamma \ \text{and}\ |S_t| \ge d_{\min},\\
\textsc{Commit}, & |S_t| \ge d_{\max},\\
\textsc{Continue}, & \text{otherwise}.
\end{cases}
\end{equation}
The peak probability \(\hat{p}_t\) is used to shape the learned representation during training and is not required at inference unless otherwise specified.

Algorithm~\ref{alg:epistream_online} summarizes the online episode formation procedure. EpiStream maintains an open episode candidate and a budgeted episodic memory bank. At each timestep, it predicts the close-versus-continue advantage; when the stopping condition is satisfied, the candidate is closed into an episodic memory unit and a new candidate begins.

\section{Experiments}
\label{sec:experiments}

In this section, we present our experimental setup, results, and analysis. We answer the following research questions (RQs):  \textbf{RQ1:} Does EpiStream form better semantic episodes for dense-action stream understanding? \textbf{RQ2:} Do EpiStream episodes improve downstream reasoning under the same memory budget? \textbf{RQ3:} Can our two-head advantage learning stably approximate offline utility-optimal episode formation?

\subsection{Experimental Setup}
\label{sec:exp_setup}

\paragraph{Dataset.}

To evaluate dense-action streaming understanding, we construct \textbf{DAS-Utility}, a benchmark comprising 60 proprietary first-person shooter (FPS) gameplay sessions (each exceeding 1200 seconds) collected for this study, spanning a total corpus of 107.3 hours.
The corpus contains approximately 153,600 sampled video frames temporally aligned with 151,163 timestamped events parsed from game-engine telemetry.
Crucially, telemetry is \textbf{strictly offline}: it provides privileged multi-modal signals, including player coordinates, viewing yaw, health status, combat interactions, damage exchanges, item acquisition, and teammate context, which are used exclusively to construct unbiased utility-optimal training targets and evaluation references. The online policy operates under strict causal constraints and never receives telemetry at inference time.

\begin{figure}[!t]
    \centering
    \includegraphics[width=0.9\linewidth]{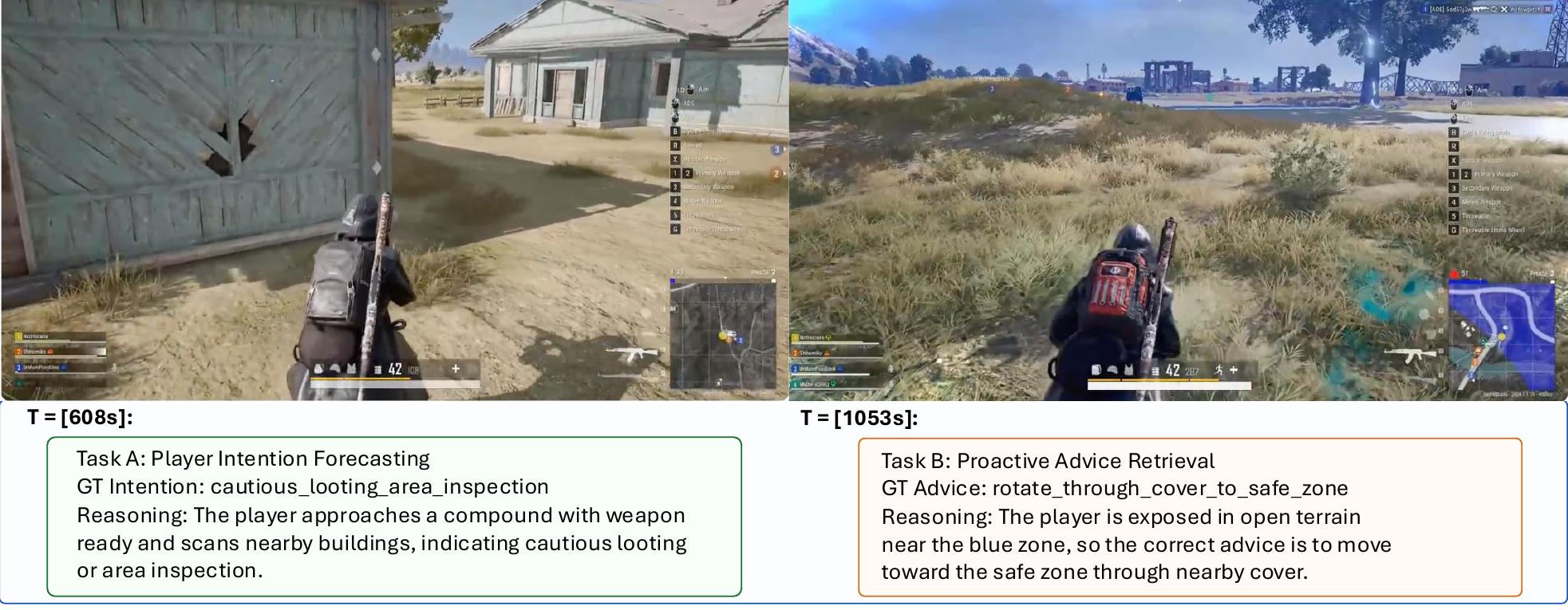}
    \caption{
        An example of downstream task evaluation result. Left: Player Intention Forecasting; Right: Proactive Advice Retrieval
        }
    \label{fig:case}
\end{figure}

As illustrated in Figure~\ref{fig:case}, reference labels for downstream tasks are constructed using offline privileged context:
\begin{itemize}[leftmargin=*, noitemsep, topsep=2pt]
\item \textbf{Player Intention Forecasting (Task A):} Reference labels describe the player's tactical state over the subsequent 15 seconds (e.g., \textit{cautious looting area inspection} in Figure~\ref{fig:case}, left).
\item \textbf{Proactive Advice Retrieval (Task B):} Reference labels specify actionable time intervals paired with ideal intervention timing and natural-language reasoning (e.g., \textit{rotate through cover to safe zone} in Figure~\ref{fig:case}, right), curated via a VLM-assisted pipeline and verified by human annotators.
\end{itemize}
Appendix~\ref{app:DAS_dataset_fill} provides exhaustive documentation on data processing, semantic ontology mapping, and annotation pipelines.


\paragraph{Abalations.}
To isolate the effect of temporal abstraction, all methods use the same episode-to-memory constructor \(G(\cdot)\), downstream LLM, and fixed memory budget. The only difference lies in how the streaming video is organized into episodic memory units. Therefore, performance differences directly reflect the quality of episode formation rather than additional visual context or larger memory capacity.

\begin{table*}[!t]
\centering
\caption{
Metric taxonomy for semantic episode formation.
}
\label{tab:metric_definition}
\resizebox{\textwidth}{!}{
\begin{tabular}{@{}lllll@{}}
\toprule
\textbf{Group} & \textbf{Metric} & \textbf{Symbol} & \textbf{Definition} & \textbf{Direction} \\
\midrule

\multirow{5}{*}{Intra-episode coherence}
& State Entropy & Ent & Randomness of state distribution inside an episode & $\downarrow$ \\
& State Purity & Pur & Ratio of the dominant state inside an episode & $\uparrow$ \\
& Transition Rate & TR & Frequency of semantic state changes inside an episode & $\downarrow$ \\
& Tag Clarity & TagCl & Clarity and confidence of episode-level semantic tags & $\uparrow$ \\
& Semantic Consistency & SemC & Aggregated score for intra-episode semantic coherence & $\uparrow$ \\

\midrule

\multirow{3}{*}{Inter-episode distinctiveness}
& State Divergence & StatDiv & State distribution difference between adjacent episodes & $\uparrow$ \\
& Evidence Separability & EviSep & Evidence-profile difference between adjacent episodes & $\uparrow$ \\
& Overall Separability & Sep & Aggregated score for inter-episode distinctiveness & $\uparrow$ \\

\midrule

\multirow{2}{*}{Closure utility}
& Marginal Boundary Utility & MBU & Utility gain of closing compared with merging episodes & $\uparrow$ \\
& Positive Closure Ratio & MBU+\% & Percentage of episode closures with positive utility & $\uparrow$ \\

\bottomrule
\end{tabular}
}
\end{table*}

\paragraph{Metrics.} 

We summarize the diagnostic metrics for semantic episode formation in Table~\ref{tab:metric_definition}. Evaluating whether a formed episode is internally coherent, distinguishable from neighboring episodes, and useful as a closure point for episodic memory construction. 

For downstream evaluation, we report Top-1 accuracy and Soft accuracy~\cite{sokolova2009systematic,imagenet_cvpr09} for
player intention forecasting, where Soft accuracy gives credit to compatible
top-3 intent labels. For proactive advice retrieval, we report WCov@K, TU\_sum,
nDCG@K, and latency~\cite{jarvelin2002cumulated,manning2008introduction}. WCov@K measures weighted coverage of ground-truth advice
regions under one-to-one matching, TU\_sum measures matched temporal utility,
nDCG@K evaluates ranking quality, and latency measures how late the retrieved
advice is relative to the ideal opportunity center.

\paragraph{Baselines.}
We compare EpiStream against both fixed-width partition strategies and recent online temporal abstraction and segmentation methods, including DisPider, OnGEBD, and EM-Garde~\cite{qian2025dispider,jung2025onlinegebd,zheng2026emgarde}. These approaches represent the most relevant prior frameworks for online video partitioning under streaming constraints, relying on query relevance, temporal boundary detection, or action-transition signals to organize long video streams. We adapt all baselines to the same episodic memory construction setting for fair comparison under identical memory and downstream reasoning budgets.

\subsection{Episode Semantic Coherence Comparison (RQ1)}
\label{sec:rq1}

\begin{table*}[!t]
    \caption{
    Diagnostic utility metrics for semantic episode formation on dense-action sessions.
    }
    \label{tab:segquality_detail}
    \centering
    \scriptsize
    \setlength{\tabcolsep}{2.4pt}
    \renewcommand{\arraystretch}{0.92}
    \resizebox{\textwidth}{!}{
    \begin{tabular}{l r c c c c c c c c c c}
      \toprule
      & & \multicolumn{5}{c}{\textit{Intra-episode coherence}}
        & \multicolumn{3}{c}{\textit{Inter-episode distinctiveness}}
        & \multicolumn{2}{c}{\textit{Closure utility}} \\
      \cmidrule(lr){3-7}
      \cmidrule(lr){8-10}
      \cmidrule(lr){11-12}
      Method & \#Ep
      & Ent $\downarrow$
      & Pur $\uparrow$
      & TR $\downarrow$
      & TagCl $\uparrow$
      & SemC $\uparrow$
      & StatDiv $\uparrow$
      & EviSep $\uparrow$
      & Sep $\uparrow$
      & wMBU $\uparrow$
      & AvgMBU $\uparrow$ \\
      \midrule
      \multicolumn{12}{l}{\textit{Fixed-width baselines}} \\
      \quad Fixed-10s
        & 140.6 & 0.199 & 0.761 & 0.135 & 0.719 & 0.736
        & 0.303 & 2.422 & 1.363 & 0.250 & 0.189 \\
      \quad Fixed-15s
        & 93.9 & 0.257 & 0.705 & 0.209 & 0.657 & 0.671
        & 0.335 & 2.479 & 1.407 & 0.282 & 0.272 \\
      \quad Fixed-30s
        & 47.2 & 0.366 & 0.606 & 0.407 & 0.549 & 0.547
        & 0.371 & 2.589 & 1.480 & 0.326 & 0.312 \\
      \midrule
      \multicolumn{12}{l}{\textit{Generic temporal segmentation}} \\
      \quad DisPider
        & 25.1 & 0.494 & 0.530 & 0.841 & 0.469 & 0.407
        & 0.407 & 1.487 & 0.947 & 0.214 & 0.191 \\
      \quad OnGEBD
        & 173.1 & 0.280 & 0.692 & 0.421 & 0.640 & 0.621
        & 0.415 & 2.657 & 1.536 & 0.340 & 0.302 \\
      \quad EM-Garde
        & 64.8 & 0.302 & 0.680 & 0.459 & 0.631 & 0.608
        & 0.394 & 1.659 & 1.026 & 0.241 & 0.229 \\
      \midrule
      \multicolumn{12}{l}{\textit{Utility-aware online policy}} \\
      \quad \textbf{EpiStream}
        & 15.5 & \textbf{0.092} & \textbf{0.896} & \textbf{0.081} & \textbf{0.878} & \textbf{0.882}
        & \textbf{0.476} & \textbf{2.776} & \textbf{1.626} & \textbf{0.385} & \textbf{0.522} \\
        \quad Rel. improv. over best baseline
        & N/A & +53.8\% & +17.7\% & +40.0\% & +22.1\% & +19.8\%
        & +14.7\% & +4.5\% & +5.9\% & +13.8\% & $+$67.3\% \\
      \bottomrule
    \end{tabular}
    }
\end{table*}

\begin{table*}[t]
\centering
\caption{
Downstream task evaluation under a fixed memory budget. Left: Player Intention Forecasting, Right: Proactive Advice Retrieval.
}
\label{tab:downstream_tasks}
\vspace{-4pt}

\begin{minipage}[t]{0.47\textwidth}
\centering{\footnotesize\textbf{Player Intention Forecasting}}\\[1pt]
\scriptsize
\setlength{\tabcolsep}{3.2pt}
\renewcommand{\arraystretch}{0.91}
\begin{tabular}{lcc}
\toprule
Method & Top-1 Acc $\uparrow$ & Soft Acc $\uparrow$ \\
\midrule
\multicolumn{3}{l}{\textit{No episodic memory}} \\
\quad No-Memory & 0.08 & 0.72 \\
\midrule
\multicolumn{3}{l}{\textit{Fixed-width baselines}} \\
\quad Fixed-10s & 0.30 & 0.84 \\
\quad Fixed-20s & 0.28 & 0.88 \\
\quad Fixed-30s & 0.22 & 0.86 \\
\midrule
\multicolumn{3}{l}{\textit{Generic online segmentation}} \\
\quad OnGEBD & 0.32 & 0.78 \\
\quad EM-Garde & 0.28 & 0.82 \\
\quad DisPider & 0.10 & 0.80 \\
\midrule
\multicolumn{3}{l}{\textit{Utility-aware episode formation}} \\
\quad \textbf{EpiStream} & \textbf{0.38} & \textbf{0.92} \\
\bottomrule
\end{tabular}
\end{minipage}%
\hfill%
\begin{minipage}[t]{0.50\textwidth}
\centering
{\footnotesize\textbf{Proactive Advice Retrieval}}\\[1pt]
\scriptsize
\setlength{\tabcolsep}{2.3pt}
\renewcommand{\arraystretch}{0.91}
\begin{tabular}{lcccc}
\toprule
Method & WCov@K $\uparrow$ & TU\_sum $\uparrow$ & nDCG@K $\uparrow$ & Lat. $\downarrow$ \\
\midrule
\multicolumn{5}{l}{\textit{No episodic memory}} \\
\quad No-Memory & 0.204 & 0.773 & N/A & 22.7 \\
\midrule
\multicolumn{5}{l}{\textit{Fixed-width baselines}} \\
\quad Fixed-10s & 0.020 & 0.101 & 0.071 & N/A \\
\quad Fixed-20s & 0.095 & 0.439 & 0.152 & N/A \\
\midrule
\multicolumn{5}{l}{\textit{Generic online segmentation}} \\
\quad OnGEBD & 0.074 & 0.330 & 0.148 & 6.7 \\
\quad EM-Garde & 0.104 & 0.492 & 0.170 & 4.6 \\
\quad DisPider & 0.425 & 1.693 & 0.494 & 17.7 \\
\midrule
\multicolumn{5}{l}{\textit{Utility-aware episode formation}} \\
\quad \textbf{EpiStream} & \textbf{0.466} & \textbf{1.855} & \textbf{0.509} & 14.4 \\
\bottomrule
\end{tabular}
\end{minipage}

\vspace{2pt}
\begin{minipage}{0.97\textwidth}
\footnotesize
The downstream LLM receives the same recent raw video and the same memory budget for all methods.
For intention forecasting, it uses the $K{=}3$ most recent episode cards and predicts the player's tactical intent over the next 15 seconds.
\end{minipage}

\vspace{-4pt}
\end{table*}

As shown in Table~\ref{tab:segquality_detail}, EpiStream consistently yields the highest-quality semantic episodes across all three utility dimensions.
Unlike heuristics or transition-based baselines that either over-segment continuous action streams or rely on naive duration priors, EpiStream forms significantly more stable and semantically purer memory units, as reflected by substantial reductions in internal state entropy and transition rate alongside higher overall separability.
Crucially, the elevated marginal boundary utility (wMBU and AvgMBU) confirms that its decision policy commits boundaries when they actively maximize downstream abstraction value rather than merely triggering on low-level visual shifts.
Furthermore, across 12 different utility reweighting configurations, the top-ranked method for each diagnostic metric remained unchanged (Appendix~\ref{app:sensitivity}), demonstrating the robustness of utility-aware episode formation for dense-action streams.

\subsection{Downstream Tasks Evaluation (RQ2)}
\label{sec:rq2}

As shown in Table~\ref{tab:downstream_tasks}, EpiStream achieves superior performance across both downstream tasks under a fixed memory budget. For player intention forecasting (Table~\ref{tab:downstream_tasks}, left), EpiStream elevates Top-1 accuracy to $0.38$ and Soft accuracy to $0.92$. The No-Memory baseline achieves $0.72$ Soft accuracy but only $0.08$ Top-1, indicating that while raw video provides coarse categorical hints, it fails to disambiguate fine-grained intents without structured episodic context. Among generic alternatives, OnGEBD yields competitive Top-1 ($0.32$) but inferior Soft accuracy ($0.78$), suggesting its over-fragmented windows isolate correct boundaries at the cost of contextual coherence. For proactive advice retrieval (Table~\ref{tab:downstream_tasks}, right), EpiStream achieves top-tier results across weighted coverage ($\text{WCov@K}=0.466$), matched temporal utility ($\text{TU\_sum}=1.855$), and ranking quality ($\text{nDCG@K}=0.509$). Compared with the strongest baseline DisPider, EpiStream yields statistically significant improvements in $\text{WCov@K}$ and $\text{nDCG@K}$ alongside reduced latency ($14.4\text{s}$ vs. $17.7\text{s}$).
In contrast, short fixed-width windows (10s, 20s) underperform even the No-Memory baseline by flooding the budget with noisy, low-quality candidates. 

Together, these results reveal complementary benefits: intention forecasting rewards intra-episode coherence for state disambiguation, whereas advice retrieval demands precise closure timing and inter-episode distinctiveness. EpiStream's consistent gains across both prove that our optimized episodic memory successfully translates intermediate semantic coherence into measurable, real world downstream utility.

Viewing the results of RQ1 and RQ2 together makes the rationale behind our method even clearer. Because real-time constraints and sparse rewards make end-to-end RLHF impractical for DAS, we design a two-stage optimization framework. Specifically, Stage 1 method improves episode quality (Table \ref{tab:segquality_detail}), while in Stage2, EpiStream translates intermediate semantic coherence into measurable, real world downstream task gains (Table \ref{tab:downstream_tasks})

\subsection{Utility-optimal Episode Learning (RQ3)}
\label{sec:rq3}

\begin{figure}[!t]
    \centering
    \includegraphics[width=\linewidth]{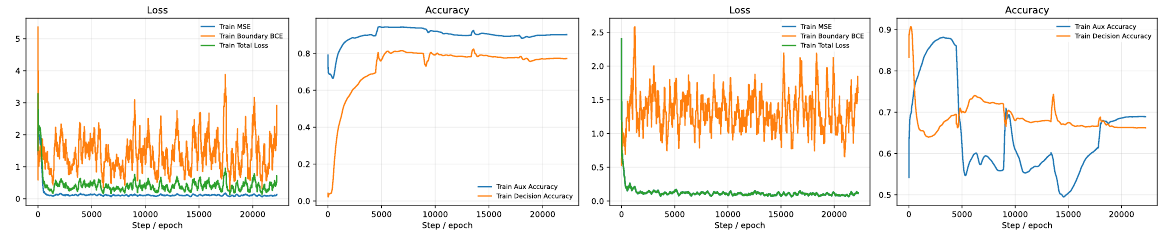}
    \caption{
    Training curves of the causal episode commitment policy. Left: EpiStream with joint advantage regression and auxiliary peak prediction. Right: policy trained without the auxiliary peak head. The joint objective stabilizes causal commit learning, improves discrimination of high-advantage \textsc{Commit} regions, and better preserves decision-critical peaks in the offline optimal.
    }
    \label{fig:learning_curves}
\end{figure}

Since high-quality episode closures manifest as sparse peaks in the advantage landscape, standard regression risks fitting the dominant \textsc{Continue} background while underweighting critical \textsc{Commit} decisions. As shown in Table~\ref{tab:objective_ablation}, while the pure regression and our joint two-head objectives achieve comparable validation MSE, the auxiliary peak-weighted head yields substantial gains in diagnostic episode quality (e.g., higher \textit{wSemC} and \textit{wSep}), confirming its necessity in localizing rare, decision-critical boundaries. Figure~\ref{fig:learning_curves} further illustrates that the joint objective stabilizes causal policy convergence and prevents the suppression of sparse closure peaks. To evaluate how closely our causal online policy approximates the privileged offline optimum, Table~\ref{tab:online_oracle_gap} compares the partitions induced by EpiStream against the offline oracle and alternative online baselines. Operating under strict causal constraints without access to future frames or telemetry during inference, EpiStream closes the majority of the performance gap to the offline oracle in terms of episodic separability (\textit{wSep}) and marginal boundary utility (\textit{wMBU}), achieving a positive closure ratio exceeding $99\%$. Rather than trivial over-segmentation that merely maximizes short term window homogeneity, EpiStream successfully distills offline utility-optimal references into a practical, decision-efficient causal policy for streaming dense-action understanding.

\begin{table}[!t]
  \caption{
  Ablation on learning objectives for causal episode formation. 
  The two-head objective combines continuous advantage regression with sparse high-advantage peak prediction.
  }
  \label{tab:objective_ablation}
  \centering
  \small
  \begin{tabular}{lccccc}
    \toprule
    Objective & Val MSE $\downarrow$ & Close F1 $\uparrow$ & wSemC $\uparrow$ & wSep $\uparrow$ & wMBU $\uparrow$ \\
    \midrule
    Advantage regression only  & 0.120 & 0.098 & 0.474 & 0.861 & 0.253 \\
    Regression + peak head + peak weighting & 0.117 & 0.104 & 0.549 & 1.079 & 0.255 \\
    Offline oracle & N/A & N/A & 0.537 & 1.366 & 0.393 \\
    \bottomrule
  \end{tabular}
\end{table}

\begin{table}[!t]
  \caption{
  Online policy approximation to offline utility-derived episode references. 
  The learned EpiStream policy is causal and uses no telemetry or future frames during inference.
  }
  \label{tab:online_oracle_gap}
  \centering
  \small
  \begin{tabular}{lccccc}
    \toprule
    Method & \#Ep & wSemC $\uparrow$ & wSep $\uparrow$ & wMBU $\uparrow$ & MBU+\% $\uparrow$ \\
    \midrule
    Fixed-15s & 93.9 & 0.592 & 0.571 & 0.044 & 68.6\% \\
    EmGarde & 64.8 & 0.541 & 0.640 & 0.103 & 78.8\% \\
    \textbf{EpiStream} & 15.5  & 0.401 & 0.918 & 0.266 & 99.2\% \\
    Offline oracle & 35.3 & 0.532 & 0.931 & 0.287 & 99.4\% \\
    \bottomrule
  \end{tabular}
\end{table}

\section{Conclusions}



We presented EpiStream, a utility-aware framework for online semantic episode formation in dense-action streams. EpiStream reframes streaming temporal abstraction as a causal commitment problem: deciding when an evolving video prefix should be closed into an episodic memory unit that preserves downstream-relevant evidence. By constructing utility-derived commit advantages from privileged offline signals and distilling them into a peak-aware causal policy, EpiStream forms more coherent and utility-bearing episodes, improving both semantic episode quality and downstream reasoning under a fixed memory budget. 
The current study focuses on FPS gameplay, where synchronized telemetry enables reliable utility supervision; nonetheless, the proposed utility-aware formulation and online commitment paradigm can be readily applied to broader dense-action domains such as autonomous driving, sports commentary, and human activity streams.



\newpage

\nocite{*}
\bibliographystyle{plain}
\bibliography{neurips_2026/Bibs/references}

@inproceedings{heilbron2015activitynet,
  title={ActivityNet: A Large-Scale Video Benchmark for Human Activity Understanding},
  author={Heilbron, Fabian Caba and Escorcia, Victor and Ghanem, Bernard and Niebles, Juan Carlos},
  booktitle={Proceedings of the IEEE Conference on Computer Vision and Pattern Recognition},
  pages={961--970},
  year={2015}
}

@inproceedings{krishna2017densecaptioning,
  title={Dense-Captioning Events in Videos},
  author={Krishna, Ranjay and Hata, Kenji and Ren, Frederic and Fei-Fei, Li and Niebles, Juan Carlos},
  booktitle={Proceedings of the IEEE International Conference on Computer Vision},
  pages={706--715},
  year={2017}
}

@inproceedings{yu2019activitynetqa,
  title={ActivityNet-QA: A Dataset for Understanding Complex Web Videos via Question Answering},
  author={Yu, Zhou and Xu, Dejing and Yu, Jun and Yu, Ting and Zhao, Zhou and Zhuang, Yueting and Tao, Dacheng},
  booktitle={Proceedings of the AAAI Conference on Artificial Intelligence},
  volume={33},
  pages={9127--9134},
  year={2019}
}

@article{yeung2018everymoment,
  title={Every Moment Counts: Dense Detailed Labeling of Actions in Complex Videos},
  author={Yeung, Serena and Russakovsky, Olga and Jin, Ning and Andriluka, Mykhaylo and Mori, Greg and Fei-Fei, Li},
  journal={International Journal of Computer Vision},
  volume={126},
  number={2--4},
  pages={375--389},
  year={2018},
  publisher={Springer}
}

@article{sardari2025refdense,
  title={Reframing Dense Action Detection (RefDense): A Paradigm Shift in Problem Solving and a Novel Optimization Strategy},
  author={Sardari, Faegheh and Mustafa, Armin and Jackson, Philip J. B. and Hilton, Adrian},
  journal={arXiv preprint arXiv:2501.18509},
  year={2025}
}

@article{maaz2023videochatgpt,
  title={Video-ChatGPT: Towards Detailed Video Understanding via Large Vision and Language Models},
  author={Maaz, Muhammad and Rasheed, Hanoona and Khan, Salman and Khan, Fahad Shahbaz},
  journal={arXiv preprint arXiv:2306.05424},
  year={2023}
}

@article{lin2023videollava,
  title={Video-LLaVA: Learning United Visual Representation by Alignment Before Projection},
  author={Lin, Bin and Ye, Yang and Zhu, Bin and Cui, Jiaxi and Ning, Munan and Jin, Peng and Yuan, Li},
  journal={arXiv preprint arXiv:2311.10122},
  year={2023}
}

@inproceedings{li2024mvbench,
  title={MVBench: A Comprehensive Multi-modal Video Understanding Benchmark},
  author={Li, Kunchang and Wang, Yali and He, Yinan and Li, Yizhuo and Wang, Yi and Wang, Limin and Qiao, Yu},
  booktitle={Proceedings of the IEEE/CVF Conference on Computer Vision and Pattern Recognition},
  year={2024}
}

@article{fu2024videomme,
  title={Video-MME: The First-Ever Comprehensive Evaluation Benchmark of Multi-modal LLMs in Video Analysis},
  author={Fu, Chaoyou and Dai, Yuhan and Luo, Yongdong and Li, Lei and Ren, Shuhuai and Zhang, Renrui and Wang, Zihan and Zhou, Chenyu and Shen, Yunhang and Zhang, Mengdan and others},
  journal={arXiv preprint arXiv:2405.21075},
  year={2024}
}

@inproceedings{mangalam2023egoschema,
  title={EgoSchema: A Diagnostic Benchmark for Very Long-form Video Language Understanding},
  author={Mangalam, Karttikeya and Akshulakov, Raiymbek and Malik, Jitendra},
  booktitle={Advances in Neural Information Processing Systems},
  year={2023}
}

@inproceedings{wu2024longvideobench,
  title={LongVideoBench: A Benchmark for Long-context Interleaved Video-Language Understanding},
  author={Wu, Haoning and Li, Dongxu and Chen, Bei and Li, Junnan},
  booktitle={Advances in Neural Information Processing Systems},
  year={2024}
}

@inproceedings{zhou2024mlvu,
  title={MLVU: Benchmarking Multi-task Long Video Understanding},
  author={Zhou, Junjie and Shu, Yan and Zhao, Bo and Wu, Boya and Liang, Zhengyang and Xiao, Shitao and Qin, Minghao and Yang, Xi and Xiong, Yongping and Zhang, Bo and Huang, Tiejun and Liu, Zheng},
  booktitle={Proceedings of the IEEE/CVF Conference on Computer Vision and Pattern Recognition},
  year={2025}
}

@inproceedings{chen2024videollmonline,
  title={VideoLLM-online: Online Video Large Language Model for Streaming Video},
  author={Chen, Joya and Lv, Zhaoyang and Wu, Shiwei and Lin, Kevin Qinghong and Song, Chenan and Gao, Difei and Liu, Jia-Wei and Gao, Ziteng and Mao, Dongxing and Shou, Mike Zheng},
  booktitle={Proceedings of the IEEE/CVF Conference on Computer Vision and Pattern Recognition},
  year={2024}
}

@inproceedings{qian2024videostreaming,
  title={Streaming Long Video Understanding with Large Language Models},
  author={Qian, Rui and Dong, Xiaoyi and Zhang, Pan and Zang, Yuhang and Ding, Shuangrui and Lin, Dahua and Wang, Jiaqi},
  booktitle={Advances in Neural Information Processing Systems},
  year={2024}
}

@article{liu2024streamchat,
  title={StreamChat: Chatting with Streaming Video},
  author={Liu, Jihao and Yu, Zhiding and Lan, Shiyi and Wang, Shihao and Fang, Rongyao and Kautz, Jan and Li, Hongsheng and Alvarez, Jose M.},
  journal={arXiv preprint arXiv:2412.08646},
  year={2024}
}

@inproceedings{qian2025dispider,
  title={Dispider: Enabling Video LLMs with Active Real-Time Interaction via Disentangled Perception, Decision, and Reaction},
  author={Qian, Rui and Ding, Shuangrui and Dong, Xiaoyi and Zhang, Pan and Zang, Yuhang and Cao, Yuhang and Lin, Dahua and Wang, Jiaqi},
  booktitle={Proceedings of the IEEE/CVF Conference on Computer Vision and Pattern Recognition},
  year={2025}
}

@inproceedings{chen2025livecc,
  title={LiveCC: Learning Video LLM with Streaming Speech Transcription at Scale},
  author={Chen, Joya and Zeng, Ziyun and Lin, Yiqi and Li, Wei and Ma, Zejun and Shou, Mike Zheng},
  booktitle={Proceedings of the IEEE/CVF Conference on Computer Vision and Pattern Recognition},
  year={2025}
}

@article{wang2025streambridge,
  title={StreamBridge: Turning Your Offline Video Large Language Model into a Proactive Streaming Assistant},
  author={Wang, Haibo and Feng, Bo and Lai, Zhengfeng and Xu, Mingze and Li, Shiyu and Ge, Weifeng and Dehghan, Afshin and Cao, Meng and Huang, Ping},
  journal={arXiv preprint arXiv:2505.05467},
  year={2025}
}

@article{xu2025streamingvlm,
  title={StreamingVLM: Real-Time Understanding for Infinite Video Streams},
  author={Xu, Ruyi and Xiao, Guangxuan and Chen, Yukang and He, Liuning and Peng, Kelly and Lu, Yao and Han, Song},
  journal={arXiv preprint arXiv:2510.09608},
  year={2025}
}

@article{lu2026vista,
  title={Vista: Scene-Aware Optimization for Streaming Video Question Answering under Post-Hoc Queries},
  author={Lu, Haocheng and Zhang, Nan and Tao, Wei and Qu, Xiaoyang and Li, Guokuan and Wan, Jiguang and Wang, Jianzong},
  journal={arXiv preprint arXiv:2602.08448},
  year={2026}
}

@article{guo2026eventvstream,
  title={Event-VStream: Event-Driven Real-Time Understanding for Long Video Streams},
  author={Guo, Zhenghui and Man, Yuanbin and Sheng, Junyuan and Lin, Bowen and Ahmed, Ahmed and Jiang, Bo and Zhang, Boyuan and Yin, Miao and Jin, Sian and Gnawal, Omprakash and Zhang, Chengming},
  journal={arXiv preprint arXiv:2601.15655},
  year={2026}
}

@article{jung2025onlinegebd,
  title={Online Generic Event Boundary Detection},
  author={Jung, Hyungrok and Kim, Daneul and Lim, Seunggyun and Son, Jeany and Choi, Jonghyun},
  journal={arXiv preprint arXiv:2510.06855},
  year={2025}
}

@article{azad2026streamready,
  title={StreamReady: Learning What to Answer and When in Long Streaming Videos},
  author={Azad, Shehreen and Vineet, Vibhav and Rawat, Yogesh Singh},
  journal={arXiv preprint arXiv:2603.08620},
  year={2026}
}

@article{zheng2026emgarde,
  title={Em-Garde: A Propose-Match Framework for Proactive Streaming Video Understanding},
  author={Zheng, Yikai and Ding, Xin and Yang, Yifan and Jiang, Shiqi and Wu, Hao and Zhang, Qianxi and Wang, Weijun and Cao, Ting and Liu, Yunxin},
  journal={arXiv preprint arXiv:2603.19054},
  year={2026}
}

@inproceedings{diko2025rewind,
  title     = {{ReWind}: Understanding Long Videos with Instructed Learnable Memory},
  author    = {Diko, Anxhelo and Wang, Tinghuai and Swaileh, Wassim and Sun, Shiyan and Patras, Ioannis},
  booktitle = {Proceedings of the IEEE/CVF Conference on Computer Vision and Pattern Recognition (CVPR)},
  month     = jun,
  year      = {2025},
  pages     = {13734--13743}
}

@article{yuan2025memvid,
  title={Memory-enhanced Retrieval Augmentation for Long Video Understanding},
  author={Yuan, Huaying and Liu, Zheng and Qin, Minghao and Qian, Hongjin and Shu, Yan and Dou, Zhicheng and Wen, Ji-Rong and Sebe, Nicu},
  journal={arXiv preprint arXiv:2503.09149},
  year={2025}
}

@article{li2025ovobench,
  title={OVO-Bench: How Far is Your Video-LLMs from Real-World Online Video Understanding?},
  author={Li, Yifei and Niu, Junbo and Miao, Ziyang and Ge, Chunjiang and Zhou, Yuanhang and He, Qihao and Dong, Xiaoyi and Duan, Haodong and Ding, Shuangrui and Qian, Rui and Zhang, Pan and Zang, Yuhang and Cao, Yuhang and He, Conghui and Wang, Jiaqi},
  journal={arXiv preprint arXiv:2501.05510},
  year={2025}
}

@article{yan2026proactvl,
  title={Proact-VL: A Proactive VideoLLM for Real-Time AI Companions},
  author={Yan, Weicai and Dai, Yuhong and Ran, Qi and Li, Haodong and Lin, Wang and Liao, Hao and Xie, Xing and Jin, Tao and Lian, Jianxun},
  journal={arXiv preprint arXiv:2603.03447},
  year={2026}
}

@article{wang2026gameplayqa,
  title={GameplayQA: A Benchmarking Framework for Decision-Dense POV-Synced Multi-Video Understanding of 3D Virtual Agents},
  author={Wang, Yunzhe and Xu, Runhui and Zheng, Kexin and Zhang, Tianyi and Kogundi, Jayavibhav Niranjan and Hans, Soham and Ustun, Volkan},
  journal={arXiv preprint arXiv:2603.24329},
  year={2026}
}

@inproceedings{zhong2024onlinetas,
  title={OnlineTAS: An Online Baseline for Temporal Action Segmentation},
  author={Zhong, Qing and Ding, Guodong and Yao, Angela},
  booktitle={Advances in Neural Information Processing Systems},
  year={2024}
}

@inproceedings{zeng2025streamforest,
  title={StreamForest: Efficient Online Video Understanding with Persistent Event Memory},
  author={Zeng, Xiangyu and Qiu, Kefan and Zhang, Qingyu and Li, Xinhao and Wang, Jing and Li, Jiaxin and Yan, Ziang and Tian, Kun and Tian, Meng and Zhao, Xinhai and Wang, Yi and Wang, Limin},
  booktitle={Advances in Neural Information Processing Systems},
  year={2025}
}

@article{sokolova2009systematic,
  title={A systematic analysis of performance measures for classification tasks},
  author={Sokolova, Marina and Lapalme, Guy},
  journal={Information Processing \& Management},
  volume={45},
  number={4},
  pages={427--437},
  year={2009},
  publisher={Elsevier}
}

@inproceedings{imagenet_cvpr09,
  title={ImageNet: A large-scale hierarchical image database},
  author={Deng, Jia and Dong, Wei and Socher, Richard and Li, Li-Jia and Li, Kai and Fei-Fei, Li},
  booktitle={Proceedings of the IEEE Conference on Computer Vision and Pattern Recognition},
  pages={248--255},
  year={2009}
}

@article{jarvelin2002cumulated,
  title={Cumulated gain-based evaluation of IR techniques},
  author={J{\"a}rvelin, Kalervo and Kek{\"a}l{\"a}inen, Jaana},
  journal={ACM Transactions on Information Systems},
  volume={20},
  number={4},
  pages={422--446},
  year={2002},
  publisher={ACM}
}

@book{manning2008introduction,
  title={Introduction to Information Retrieval},
  author={Manning, Christopher D. and Raghavan, Prabhakar and Sch{\"u}tze, Hinrich},
  year={2008},
  publisher={Cambridge University Press}
}

@article{wang2025videoem,
  title={Video-EM: Event-Centric Episodic Memory for Long-Form Video Understanding},
  author={Wang, Yun and Zhang, Long and Liu, Jingren and Yan, Jiaqi and Zhang, Zhanjie and Zheng, Jiahao and Ma, Ao and Ling, Run and Yang, Xun and Wu, Dapeng and Chen, Xiangyu and Li, Xuelong},
  journal={arXiv preprint arXiv:2508.09486},
  year={2025}
}

@inproceedings{zuo2025videolucy,
  title={VideoLucy: Deep Memory Backtracking for Long Video Understanding},
  author={Zuo, Jialong and Deng, Yongtai and Kong, Lingdong and Yang, Jingkang and Jin, Rui and Zhang, Yiwei and Sang, Nong and Pan, Liang and Liu, Ziwei and Gao, Changxin},
  booktitle={Advances in Neural Information Processing Systems},
  year={2025}
}

@article{yin2025videoarm,
  title={VideoARM: Agentic Reasoning over Hierarchical Memory for Long-Form Video Understanding},
  author={Yin, Yufei and Meng, Qianke and Chen, Minghao and Ding, Jiajun and Shao, Zhenwei and Yu, Zhou},
  journal={arXiv preprint arXiv:2512.12360},
  year={2025}
}

@inproceedings{azad2025hierarq,
  title={HierarQ: Task-Aware Hierarchical Q-Former for Enhanced Video Understanding},
  author={Azad, Shehreen and Vineet, Vibhav and Rawat, Yogesh Singh},
  booktitle={Proceedings of the IEEE/CVF Conference on Computer Vision and Pattern Recognition},
  year={2025}
}







\newpage
\appendix

\section{Semantic Score Metrics}
\label{app:semantic_score_metrics}

This section describes the diagnostic metrics used to evaluate the semantic
quality of a produced episode partition. These metrics are computed from the
privileged telemetry-derived time bins and are used only for offline utility
construction and evaluation.

\paragraph{Run-based semantic consistency.}
In implementation, each time bin is assigned a dominant semantic state, and
the resulting state sequence is compressed into contiguous runs. This run-based
view avoids the length bias of bin-averaged state histograms: a long window
containing one dominant state and several short interruptions should still be
recognized as temporally fragmented. Let \(q_W(z)\) be the run-based state
distribution defined in, we compute
normalized entropy \(H(W)\), purity \(P(W)\), and transition rate
\(\mathrm{TR}(W)\) on this run distribution.

The implementation additionally uses a normalized top-label margin
\begin{equation}
\mathrm{MN}(W)
=
\frac{q_W^{(1)}-q_W^{(2)}}{q_W^{(1)}+\epsilon},
\end{equation}
where \(q_W^{(1)}\) and \(q_W^{(2)}\) are the largest and second-largest
state probabilities under \(q_W\). The reported semantic consistency score is
\begin{equation}
\mathrm{SemC}(W)
=
\frac{1}{3}\left(1-H(W)\right)
+\frac{1}{6}P(W)
+\frac{1}{6}\left(1-\mathrm{TR}(W)\right)
+\frac{1}{3}\mathrm{MN}(W).
\end{equation}
This is the score reported as SemC or wSemC in the experiments. The weighted
version wSemC is the duration-weighted average over all produced episodes:
\begin{equation}
\mathrm{wSemC}
=
\frac{\sum_i |S_i|\,\mathrm{SemC}(S_i)}
{\sum_i |S_i|}.
\end{equation}

\paragraph{Tag clarity.}
We also report tag clarity as an auxiliary diagnostic of whether an episode can
be assigned an unambiguous semantic label. It combines three run-based terms:
\begin{equation}
\mathrm{TagCl}(W)
=
\frac{1}{3}\mathrm{MN}(W)
+\frac{1}{3}P(W)
+\frac{1}{3}\left(1-H(W)\right).
\end{equation}
Tag clarity is not used as an online input; it is computed from privileged
offline state annotations for evaluation and utility construction.

\paragraph{Inter-episode distinctiveness.}
For adjacent episodes \(S_i\) and \(S_{i+1}\), we compute state divergence and
evidence separability as described in Appendix~\ref{app:evaluation_metrics_fill}. The
reported separability metric is
\begin{equation}
\mathrm{Sep}(S_i,S_{i+1})
=
\frac{1}{2}D_{\mathrm{state}}(S_i,S_{i+1})
+\frac{1}{2}D_{\mathrm{ev}}(S_i,S_{i+1}).
\end{equation}
The duration-weighted version wSep weights each boundary by the mean duration
of its two adjacent episodes:
\begin{equation}
\mathrm{wSep}
=
\frac{\sum_i \bar{d}_i\,\mathrm{Sep}(S_i,S_{i+1})}
{\sum_i \bar{d}_i},
\qquad
\bar{d}_i=\frac{|S_i|+|S_{i+1}|}{2}.
\end{equation}

\paragraph{Closure utility.}
For each predicted boundary, we compute marginal boundary utility
\(\mathrm{MBU}(S_i,S_{i+1})\). We report both the duration-weighted mean
\begin{equation}
\mathrm{wMBU}
=
\frac{\sum_i \bar{d}_i\,\mathrm{MBU}(S_i,S_{i+1})}
{\sum_i \bar{d}_i},
\end{equation}
and the positive closure ratio
\begin{equation}
\mathrm{MBU+}
=
\frac{1}{N-1}
\sum_{i=1}^{N-1}
\mathbb{I}\!\left[\mathrm{MBU}(S_i,S_{i+1})>0\right].
\end{equation}
\begin{table}[t]
\centering
\small
\begin{tcolorbox}[
    width=0.92\linewidth,
    colback=gray!8,
    colframe=black,
    boxrule=0.5pt,
    arc=0pt,
    outer arc=0pt,
    left=10pt,
    right=10pt,
    top=8pt,
    bottom=8pt,
    boxsep=0pt
]
\raggedright
\setlist[enumerate]{
    label=(\arabic*),
    leftmargin=1.8em,
    itemsep=0pt,
    topsep=2pt,
    parsep=0pt,
    partopsep=0pt
}
\setlength{\parskip}{0.65em}

\textbf{\textcolor{blue}{Player State Events}}:
\begin{enumerate}
    \item ``[0.4] Player state: position=(720108.0,868055.0,150000.0), yaw=220.8'',
    \item ``[0.4] Player state: in aircraft, health=100.0, third-person view'',
    \item ``[39.1] Player state: parachuting, health=100.0, third-person view'',
    \item ``[124.9] Player state: standing, health=100.0, weapon=P1911, ammo=7/0'',
\end{enumerate}

\textbf{\textcolor{blue}{Item Pickup Events}}:
\begin{enumerate}
    \item ``[88.8] Item pickup: item=530001, count=1, source=ground'',
    \item ``[103.4] Item pickup: backpack(level 1), count=1, source=ground'',
    \item ``[124.7] Item pickup: P1911, count=1, source=ground'',
    \item ``[139.7] Item pickup: energy drink, count=1, source=ground'',
    \item ``[145.0] Item pickup: bandage, count=5, source=ground'',
\end{enumerate}

\textbf{\textcolor{blue}{Building Events}}:
\begin{enumerate}
    \item ``[124.4] Enter building'',
    \item ``[129.0] Leave building'',
\end{enumerate}

\textbf{\textcolor{blue}{Combat Events}}:
\begin{enumerate}
    \item ``[180.7] Fire weapon: P1911, shots=1, hits=0'',
    \item ``[180.8] Damage received: 2.3, current health=97.7, attacker=2563(AI)'',
    \item ``[181.2] Attack player: 2563(AI), weapon=P1911, damage=30.3'',
    \item ``[181.4] Fire weapon: P1911, shots=3, hits=1'',
    \item ``[181.8] Knock down player: 2563(AI), weapon=P1911, distance=1353.6'',
\end{enumerate}

\textbf{\textcolor{blue}{Visibility Events}}:
\begin{enumerate}
    \item ``[132.4] One player enters view: 2563(AI)'',
    \item ``[134.4] One player leaves view: 2563(AI)'',
\end{enumerate}

\end{tcolorbox}
\caption{Shortened example of the raw telemetry stream used for offline construction.}
\label{tab:telemetry}
\end{table}
\section{Offline Utility Construction and Online Policy Learning}
\label{app:offline_utility_construction}

Offline utility construction converts privileged telemetry into dense
supervision for a causal online policy. The procedure has four stages.

\paragraph{Telemetry binning.}
Each gameplay session is parsed into a sequence of timestamped telemetry
events. Events are mapped into a fixed semantic ontology
\[
\mathcal{Z}=\{\textsc{Parachute},\textsc{Looting},\textsc{Combat},
\textsc{Rotate},\textsc{Observe},\textsc{Recover},\textsc{Unknown}\}.
\]
We aggregate events into one-second bins. Each bin stores the active semantic
states, a dominant state, and a privileged evidence vector containing movement,
combat, damage, health, looting, visibility, and team-context signals.

\paragraph{Candidate episode references.}
The offline segmenter first proposes candidate boundaries from privileged
state changes and evidence shifts. Candidate partitions are then evaluated by
the utility metrics above. A boundary is retained when keeping it yields
positive marginal utility relative to merging its two neighboring episodes.
Thus, the offline reference is not a manually annotated visual boundary; it is
a utility-derived episode partition.

\paragraph{Dense advantage targets.}
For each sampled online decision state at time \(t\), we compare the utility
of closing the current open episode with the utility of continuing it:
\begin{equation}
A_t
=
U_{\mathrm{close}}(t)-U_{\mathrm{continue}}(t).
\end{equation}
The prepared training cache stores both normalized utilities and their
difference:
\begin{equation}
\texttt{advantage}
=
\texttt{u\_close}-\texttt{u\_continue}.
\end{equation}
The online model is trained to predict this scalar advantage from causal
video-prefix observations only.

\paragraph{Sampling around sparse utility peaks.}
Since high-advantage closure points are sparse, the dataset construction uses
denser temporal sampling around reference boundary regions and sparser sampling
in stable regions. In our main runs, the sampling interval is \(1.0\)s near
boundaries and \(2.0\)s in stable regions, with an \(8.0\)s boundary window.
Each sample contains the current memory context, the current open episode
description, a short video clip of \(8\) frames, and the offline advantage
target.

\paragraph{Online policy prompt example.}
At inference time, the policy receives only the causal memory state, the
current open episode, and the current video frames. A shortened prompt example
is shown below:
The prompt contains only information available from the streaming video and
previously committed memory. It does not include telemetry fields such as
ground-truth state labels, utility values, future events, or offline oracle
boundaries.

\paragraph{Memory schema and prompt format.}
Each committed episode is converted into a structured memory card. The memory
card stores the temporal span, the dominant episode type, a concise summary,
and task-relevant evidence:

\begin{table}[t]
\centering
\small
\begin{tcolorbox}[
    width=0.92\linewidth,
    colback=gray!8,
    colframe=black,
    boxrule=0.5pt,
    arc=0pt,
    outer arc=0pt,
    left=10pt,
    right=10pt,
    top=8pt,
    bottom=8pt,
    boxsep=0pt
]
\raggedright
\setlength{\parskip}{0.6em}
\setlist[enumerate]{
    label=(\arabic*),
    leftmargin=1.8em,
    itemsep=3pt,
    topsep=2pt,
    parsep=0pt,
    partopsep=0pt
}
\setlist[itemize]{
    label={--},
    leftmargin=1.8em,
    itemsep=1pt,
    topsep=1pt,
    parsep=0pt,
    partopsep=0pt
}

\textbf{\textcolor{blue}{Committed Memory Cards}}:
\begin{enumerate}
    \item \textit{Time:} 0.0s--86.0s $\vert$ \textit{Type:} Parachute \\
    \textit{Summary:} The player stayed in the aircraft, parachuted, and approached the landing area. \\
    \textit{Evidence:} Player remained airborne during the opening phase; No combat or looting evidence was observed.

    \item \textit{Time:} 86.0s--124.0s $\vert$ \textit{Type:} Looting \\
    \textit{Summary:} The player landed and began collecting early resources near buildings. \\
    \textit{Evidence:} Landed after parachuting; Picked up initial equipment; Moved through the first looting area.
\end{enumerate}

\textbf{\textcolor{blue}{Current Window}}: \par
\textit{Start:} 124.0s $\vert$
\textit{Current time:} 181.8s $\vert$
\textit{Duration:} 57.8s \par
\textit{Tentative type:} Combat / Looting \par
\textit{Summary so far:} The player entered and left buildings, picked up a P1911, energy drink, and bandages, then encountered an enemy and exchanged pistol fire. \par
\textit{Evidence so far:}
\begin{itemize}
    \item Entered building at 124.4s and picked up P1911.
    \item Enemy appeared in view around 132.4s.
    \item Player fired P1911 and received damage.
    \item Player knocked down the enemy at 181.8s.
\end{itemize}

\textbf{\textcolor{blue}{Current Video}}: \par
$\langle$8 sampled gameplay frames from the current open episode$\rangle$

\textbf{\textcolor{blue}{Question}}: \par
Should CURRENT\_WINDOW be committed as a complete memory card now? \par
Output exactly one token: \texttt{<CLOSE>} or \texttt{<CONTINUE>}

\end{tcolorbox}
\caption{Shortened memory prompt used by the online policy.}
\label{tab:memory_prompt}
\end{table}

\paragraph{Loss Function}
\label{app:loss_function}

The implementation follows the objective in
Section~\ref{sec:policy_learning}
. For each sample \(t\), the target
advantage is
\[
A_t=u_{\mathrm{close}}(t)-u_{\mathrm{continue}}(t).
\]
The auxiliary target is derived from the same advantage curve:
\begin{equation}
b_t=\mathbb{I}\!\left[A_t\ge \eta_{\mathrm{aux}}\right].
\end{equation}
In the main setting, \(\eta_{\mathrm{aux}}=-0.05\). The peak-weighted
regression weight is
\begin{equation}
w_t=
\begin{cases}
w_{\mathrm{peak}}, & A_t\ge \eta_{\mathrm{peak}},\\
1, & \text{otherwise},
\end{cases}
\end{equation}
where \(w_{\mathrm{peak}}=2.0\) and \(\eta_{\mathrm{peak}}=-0.10\) in the
0428 run.

The regression loss is
\begin{equation}
\mathcal{L}_{\mathrm{adv}}
=
\frac{1}{|\mathcal{T}|}
\sum_{t\in\mathcal{T}}
w_t(\hat{A}_t-A_t)^2.
\end{equation}
The auxiliary peak loss is binary cross-entropy with optional positive-class
reweighting:
\begin{equation}
\mathcal{L}_{\mathrm{peak}}
=
-\frac{1}{|\mathcal{T}|}
\sum_{t\in\mathcal{T}}
\left[
\alpha b_t\log \hat{p}_t
+(1-b_t)\log(1-\hat{p}_t)
\right].
\end{equation}
The final loss is
\begin{equation}
\mathcal{L}
=
\mathcal{L}_{\mathrm{adv}}
+\lambda_{\mathrm{peak}}\mathcal{L}_{\mathrm{peak}},
\end{equation}
with \(\lambda_{\mathrm{peak}}=0.2\) in the main two-head model. The
regression-only ablation sets \(\lambda_{\mathrm{peak}}=0\), while keeping the
same dataset and advantage targets.

\paragraph{Probabilistic interpretation of the loss terms.}
Under a Gaussian observation model with mean \(\hat A_t\) and
fixed variance \(\sigma^2\), the per-sample negative log-likelihood is
\[
-\log p(A_t\mid h_t)
= \frac{(A_t-\hat A_t)^2}{2\sigma^2} + \mathrm{const},
\]
which gives the squared-error term up to scale and an additive constant.
For the auxiliary target \(b_t\), a Bernoulli model with parameter
\(\hat p_t\) gives
\[
-\log p(b_t\mid h_t)
= -b_t\log\hat p_t - (1-b_t)\log(1-\hat p_t),
\]
which is the binary cross-entropy term. Our implementation reweights
these base losses and combines them as specified above. Because \(b_t\)
is derived from \(A_t\), the two terms provide complementary supervision
for advantage magnitude and high-advantage region membership.

\begin{table}[t]
\centering
\small
\begin{tcolorbox}[
    width=0.92\linewidth,
    colback=gray!8,
    colframe=black,
    boxrule=0.5pt,
    arc=0pt,
    outer arc=0pt,
    left=10pt,
    right=10pt,
    top=8pt,
    bottom=8pt,
    boxsep=0pt
]
\raggedright
\setlength{\parskip}{0.35em}
\setlist[description]{
    leftmargin=2.2em,
    labelindent=0pt,
    labelsep=0.6em,
    itemsep=2pt,
    topsep=2pt,
    parsep=0pt,
    partopsep=0pt,
    font=\normalfont\ttfamily
}

\textbf{\textcolor{blue}{Memory Card Schema}}:
\begin{description}
    \item[start\_time] float,
    \item[end\_time] float,
    \item[duration] float,
    \item[episode\_type] one of [
        ``Parachute'', ``Looting'', ``Combat'', ``Rotate'',
        ``Observe'', ``Recover'', ``Unknown''
    ],
    \item[summary] string,
    \item[key\_evidence] [string, ...],
    \item[utility\_score] optional float,
\end{description}

\end{tcolorbox}
\caption{Schema of a memory card.}
\label{tab:memory_card_schema}
\end{table}

\paragraph{Training setup.}
We train the online episode-commitment policy using offline advantage targets
constructed from privileged telemetry. The policy is initialized from
Qwen2.5-VL-3B-Instruct and adapted with LoRA, while lightweight regression and
auxiliary peak-classification heads are trained on top of the shared VLM
representation. The model observes only causal inputs: previous memory cards,
the current open episode description, sampled video frames from the candidate
window, and a boundary query.

All training runs are conducted on a single NVIDIA H20 GPU. We use batch size
1 for 5 epochs and optimize with AdamW using learning rate \(2\times10^{-5}\).
Each candidate window samples 8 video frames. The main objective is
peak-weighted advantage regression, combined with an auxiliary BCE loss for
high-advantage regions. We set the auxiliary loss weight to \(0.2\), the
auxiliary advantage threshold to \(-0.05\), the peak-regression threshold to
\(-0.10\), and the peak MSE weight to \(2.0\). During online inference, the
policy commits an episode when the predicted advantage exceeds
\(\gamma=-0.47\), subject to the minimum and maximum duration constraints.

\section{Details of DAS-Utility Dataset}
\label{app:DAS_dataset_fill}

DAS-Utility is built from long-form first-person competitive gameplay streams
paired with synchronized telemetry logs. Each session is longer than 1200s and
contains dense, overlapping actions. The video stream supplies the causal
online observation, while the telemetry supplies privileged supervision for
offline utility construction.

\paragraph{Raw sources.}
For each match, the raw data include a gameplay video, extracted video frames,
and a millisecond-level event log. The event log contains player position,
viewing direction, health, in-air/on-ground state, item acquisition, combat and
damage events, enemy exposure, and teammate context. These signals are aligned
to video time and converted into one-second semantic bins.
\paragraph{Semantic ontology.}
Telemetry events are mapped into a compact set of high-level gameplay states:
\textsc{Parachute}, \textsc{Looting}, \textsc{Combat}, \textsc{Rotate},
\textsc{Observe}, \textsc{Recover}, and \textsc{Unknown}. A bin may contain
multiple low-level evidence signals, but it receives one dominant state for
run-based metric computation.

\paragraph{Dataset samples.}
Each training sample corresponds to one causal decision point. The prompt
contains: previously committed memory, the current open episode summary, a
short natural-language query asking whether to close or continue, and the
current video frames. The learning target is not a hard human boundary label;
it is the offline advantage value \(A_t\). The hard boundary indicator is kept
only for monitoring decision accuracy.

\paragraph{Causal restriction.}
At inference time, the policy observes only the current video prefix, the
current open episode, and previously committed memory. Future frames and
telemetry-derived privileged signals are unavailable.

\subsection{Dataset Construction and Annotation Pipeline}
\label{app:dataset_construction}

\paragraph{VLM-assisted candidate generation.}
To construct high-quality advice-timing annotations, we use a
VLM-assisted annotation pipeline followed by human verification. Given a full
gameplay session, we first parse the synchronized telemetry log into a
time-ordered textual event stream. The telemetry contains privileged signals
such as player position, health changes, combat events, item acquisition,
teammate status, visibility cues, and movement phases. These signals are used
only for offline annotation and utility construction; they are never provided
to the online policy at inference time.

We use Gemini 3.1 Pro as the annotation assistant. For each match and each
tactical advice query, Gemini receives the complete telemetry log and is asked
to identify a set of valid advice opportunities. Each opportunity is required
to be a temporal region rather than a single timestamp, with a center time
indicating the ideal advice timing. This region-based format reduces ambiguity:
an advice opportunity is often valid over a short interval, while the center
time represents the most actionable moment inside that interval.

The prompt used for VLM-assisted generation follows the template in
\texttt{src/advice\_timing\_retrieval\_refined.py}. The system prompt asks the
model to act as a PUBG coaching analyst and output non-redundant advice
opportunities in JSON format:
\[
\left\{
\texttt{span\_start},
\texttt{span\_end},
\texttt{center\_time},
\texttt{weight},
\texttt{reason}
\right\}.
\]
The user prompt specifies the match duration, the advice type, the natural
language description of that advice type, relevant telemetry signals to look
for, and the complete match telemetry log.

\paragraph{Prompt template for advice opportunity generation.}
We use the following prompt structure for VLM-assisted candidate generation:

\begin{table}[t]
\centering
\caption{Prompt template for VLM-assisted advice opportunity candidate generation. The VLM receives a full telemetry log and a tactical query, then outputs structured opportunity annotations.}
\small
\renewcommand{\arraystretch}{0.85}
\begin{tabular}{p{0.95\textwidth}}
\toprule
\multicolumn{1}{c}{\textbf{Advice Opportunity Generation Protocol}} \\
\midrule
\textcolor{blue}{\textbf{System Role}} \\[2pt]
You are a PUBG coaching analyst. You will receive the complete telemetry log of one match. Your task is to identify a set of valid \textbf{advice opportunities} for one tactical advice query. \\[6pt]

\textcolor{blue}{\textbf{Input}} \\[2pt]
\quad \textcolor{orange}{\textbf{Telemetry Log}} \; {\scriptsize (full match event stream, cf.\ Table~\ref{tab:telemetry})} \\
\quad \textcolor{orange}{\textbf{Tactical Query}} \; {\scriptsize (e.g., ``When should the player rotate to the safe zone?'')} \\[6pt]

\textcolor{blue}{\textbf{Constraints}} \\[2pt]
(1) A query can have multiple valid opportunities. \\
(2) Each opportunity must be a \emph{time region}, not a single second. \\
(3) The region captures when the advice is valid; the \texttt{center\_time} captures the ideal moment within that region. \\
(4) Opportunities must be semantically distinct and non-redundant. \\[6pt]

\textcolor{blue}{\textbf{Output}} \quad {\scriptsize (one JSON object per opportunity)} \\[2pt]
\quad \texttt{\{"span\_start": $t_s$, "span\_end": $t_e$, "center\_time": $t_c$, "weight": $w$, "reason": $r$\}} \\[3pt]

\midrule
{\scriptsize \textbf{Fields:} \texttt{span\_start}/\texttt{span\_end} define the valid window (seconds); \texttt{center\_time} $\in [t_s, t_e]$ is the ideal trigger point; \texttt{weight} $\in (0,1]$ reflects opportunity importance; \texttt{reason} is a free-form justification string.} \\
\bottomrule
\end{tabular}
\renewcommand{\arraystretch}{1.0}
\label{tab:advice_prompt}
\end{table}

\paragraph{Human filtering and refinement.}
The VLM-generated candidates are treated as annotation proposals rather than
final ground truth. Human annotators then review each proposed advice
opportunity using both the gameplay video and the telemetry context. During
this stage, annotators remove false positives, merge redundant opportunities,
adjust temporal spans and center times, and refine the natural-language
reasoning when necessary. Candidates are retained only when the advice is
actionable, temporally well grounded, and supported by visible gameplay or
telemetry evidence.

This human-in-the-loop process gives us dense but reliable advice-timing
references. The VLM improves recall by surfacing candidate moments across long
matches, while human verification controls precision and temporal alignment.
The resulting annotations are used as offline references for downstream
advice-timing evaluation, but not as inputs to the learned online segmentation
policy.

\paragraph{Advice opportunity schema.}
Each curated advice opportunity is stored as
\[
g_j =
\left(
s_j,
e_j,
c_j,
w_j,
r_j
\right),
\]
where \(s_j\) and \(e_j\) are the start and end times of the valid advice
region, \(c_j\) is the ideal center time, \(w_j\) is an importance weight, and
\(r_j\) is a concise human-readable reason. In some annotations, we also keep
visual or telemetry evidence fields, such as UI cues, player-state cues, and
team-context cues, to support later inspection and error analysis.

\section{Evaluation Metrics}
\label{app:evaluation_metrics_fill}

We use two classes of metrics: diagnostic episode-quality metrics and
downstream reasoning metrics.

\paragraph{Notation.}
For a produced partition \(\mathcal{S}=\{S_i\}_{i=1}^{N}\), each episode
\(S_i\) consists of time bins. Let \(\mathcal{Z}\) denote the semantic state
set. For each bin \(t\), \(z_t\) is its dominant semantic state and
\(\mathbf{e}_t\in\mathbb{R}^{|\mathcal{E}|}\) is its privileged evidence
vector. Privileged states and evidence are used only for offline evaluation.

\paragraph{Run-based intra-episode metrics.}
For each episode \(S_i\), we convert its dominant-state sequence into
contiguous runs
\[
\mathcal{R}(S_i)=\{(z_r,\ell_r)\}_{r=1}^{R_i},
\]
where \(z_r\) is the state label of run \(r\), \(\ell_r\) is its duration,
and \(R_i\) is the number of runs. The run-based state distribution is
\[
q_i(z)=\frac{1}{R_i}\sum_{r=1}^{R_i}\mathbb{I}[z_r=z].
\]

The normalized run entropy is
\[
\mathrm{Ent}(S_i)
=
-\frac{1}{\log|\mathcal{Z}|}
\sum_{z\in\mathcal{Z}} q_i(z)\log q_i(z).
\]

Run purity is
\[
\mathrm{Pur}(S_i)=\max_{z\in\mathcal{Z}}q_i(z).
\]

The normalized transition rate is
\[
\mathrm{TR}(S_i)
=
\frac{\min(R_i-1,|\mathcal{Z}|)}{|\mathcal{Z}|}.
\]

Let \(q_i^{(1)}\) and \(q_i^{(2)}\) denote the largest and second largest
values of \(q_i(z)\). The normalized margin is
\[
\mathrm{MN}(S_i)
=
\frac{q_i^{(1)}-q_i^{(2)}}{q_i^{(1)}+\epsilon}.
\]

The raw semantic consistency score is
\[
\mathrm{SemC}_{\mathrm{raw}}(S_i)
=
\frac{1}{3}\left(1-\mathrm{Ent}(S_i)\right)
+
\frac{1}{6}\mathrm{Pur}(S_i)
+
\frac{1}{6}\left(1-\mathrm{TR}(S_i)\right)
+
\frac{1}{3}\mathrm{MN}(S_i).
\]

Following the implementation, we apply a length-aware correction to avoid
over-rewarding very short single-run episodes. Let \(T_i=|S_i|\) be the
number of bins and \(\tau=9\). The reported semantic consistency is
\[
\mathrm{SemC}(S_i)
=
\begin{cases}
\mathrm{SemC}_{\mathrm{raw}}(S_i), & R_i\ge 2,\\
\left(1-\exp(-T_i/\tau)\right)\mathrm{SemC}_{\mathrm{raw}}(S_i), & R_i=1.
\end{cases}
\]

The duration-weighted semantic consistency reported as wSemC is
\[
\mathrm{wSemC}
=
\frac{\sum_{i=1}^{N}|S_i|\mathrm{SemC}(S_i)}
{\sum_{i=1}^{N}|S_i|}.
\]

Tag clarity is retained as a diagnostic score:
\[
\mathrm{TagCl}(S_i)
=
\frac{1}{3}\mathrm{MN}(S_i)
+
\frac{1}{3}\mathrm{Pur}(S_i)
+
\frac{1}{3}\left(1-\mathrm{Ent}(S_i)\right).
\]
In our v3 evaluation, TagCl is not used as an independent utility term because
its components are largely absorbed by \(\mathrm{SemC}\).

\paragraph{Inter-episode separability.}
For state separability, we use the bin-level state distribution
\[
p_i(z)=\frac{1}{|S_i|}\sum_{t\in S_i}\mathbb{I}[z_t=z],
\]
and compute total variation distance between adjacent episodes:
\[
\mathrm{StatDiv}(S_i,S_{i+1})
=
\frac{1}{2}
\sum_{z\in\mathcal{Z}}
\left|p_i(z)-p_{i+1}(z)\right|.
\]

For evidence separability, the implementation first computes a session-level
scale for each evidence channel \(k\):
\[
s_k=\max\left(
\operatorname{P90}\left(\{e_{t,k}:t\in X\}\right),
10^{-3}
\right).
\]
The length-normalized evidence rate of episode \(S_i\) is
\[
r_{i,k}=\frac{1}{|S_i|}\sum_{t\in S_i}e_{t,k}.
\]
The normalized evidence separability is
\[
\mathrm{EviSep}(S_i,S_{i+1})
=
\min\left(
\frac{1}{|\mathcal{E}|}
\sum_{k\in\mathcal{E}}
\frac{|r_{i,k}-r_{i+1,k}|}{s_k},
1
\right).
\]

The combined separability score is
\[
\mathrm{Sep}(S_i,S_{i+1})
=
\min\left(
\frac{1}{2}\mathrm{StatDiv}(S_i,S_{i+1})
+
\frac{1}{2}\mathrm{EviSep}(S_i,S_{i+1}),
1
\right).
\]

Consistent with the code, inter-episode separability is computed only for
adjacent episode pairs whose dominant states differ. Let
\[
\mathcal{B}_{\mathrm{cross}}
=
\{i: \operatorname{dom}(S_i)\neq \operatorname{dom}(S_{i+1})\}.
\]
The reported duration-weighted separability is
\[
\mathrm{wSep}
=
\frac{
\sum_{i\in\mathcal{B}_{\mathrm{cross}}}
\frac{|S_i|+|S_{i+1}|}{2}
\mathrm{Sep}(S_i,S_{i+1})
}{
\sum_{i\in\mathcal{B}_{\mathrm{cross}}}
\frac{|S_i|+|S_{i+1}|}{2}
}.
\]

\paragraph{Marginal boundary utility.}
For an adjacent pair \((S_i,S_{i+1})\), the utility of keeping the boundary is
\[
U_{\mathrm{cut}}(S_i,S_{i+1})
=
\alpha\mathrm{SemC}(S_i)
+
\beta\mathrm{SemC}(S_{i+1})
+
\gamma\mathrm{Sep}(S_i,S_{i+1})
-
\delta P_{\mathrm{cut}}(S_i,S_{i+1}),
\]
where \(\alpha=0.4\), \(\beta=0.4\), \(\gamma=0.2\), and \(\delta=0.1\).

The cut penalty is
\[
\begin{aligned}
P_{\mathrm{cut}}(S_i,S_{i+1})
={}& P_{\mathrm{short}}(S_i)
   + P_{\mathrm{short}}(S_{i+1}) \\
 &+ P_{\mathrm{long}}(S_i)
   + P_{\mathrm{long}}(S_{i+1}) \\
 &+ 0.3P_{\mathrm{overlap}}(S_i)
   + 0.3P_{\mathrm{overlap}}(S_{i+1}).
\end{aligned}
\]

The short-window, long-window, and overlap penalties are
\[
P_{\mathrm{short}}(S)
=
\max\left(0,\frac{d_{\min}-|S|}{d_{\min}}\right),
\]
\[
P_{\mathrm{long}}(S)
=
\min\left(
\max\left(0,\frac{|S|-d_{\max}}{d_{\max}}\right),
1
\right),
\]
\[
P_{\mathrm{overlap}}(S)
=
\frac{1}{|S|}
\sum_{t\in S}\mathbb{I}\left[|\mathcal{Z}_t|>1\right],
\]
where \(d_{\min}=5\) seconds and \(d_{\max}=180\) seconds in the default
evaluation.

If the boundary is removed, the merged utility is
\[
U_{\mathrm{merge}}(S_i\oplus S_{i+1})
=
\mathrm{SemC}(S_i\oplus S_{i+1})
-
\delta P_{\mathrm{merge}}(S_i\oplus S_{i+1}),
\]
with
\[
P_{\mathrm{merge}}(S)
=
P_{\mathrm{short}}(S)
+
P_{\mathrm{long}}(S)
+
0.3P_{\mathrm{overlap}}(S).
\]

The marginal boundary utility is
\[
\mathrm{MBU}(S_i,S_{i+1})
=
U_{\mathrm{cut}}(S_i,S_{i+1})
-
U_{\mathrm{merge}}(S_i\oplus S_{i+1}).
\]

The reported wMBU is an equal-weight average over cross-state boundaries:
\[
\mathrm{wMBU}
=
\frac{1}{|\mathcal{B}_{\mathrm{cross}}|}
\sum_{i\in\mathcal{B}_{\mathrm{cross}}}
\mathrm{MBU}(S_i,S_{i+1}).
\]
The positive closure ratio is
\[
\mathrm{MBU{+}\%}
=
\frac{1}{|\mathcal{B}_{\mathrm{cross}}|}
\sum_{i\in\mathcal{B}_{\mathrm{cross}}}
\mathbb{I}
\left[
\mathrm{MBU}(S_i,S_{i+1})>0
\right].
\]

\paragraph{Learning diagnostics.}
For online policy learning, each sampled timestep \(t\) has an offline
advantage target \(A_t\), and the model predicts \(\hat{A}_t\). The validation
MSE is
\[
\mathrm{MSE}
=
\frac{1}{|\mathcal{T}|}
\sum_{t\in\mathcal{T}}
w_t(\hat{A}_t-A_t)^2,
\]
where \(w_t=1\) unless peak-weighted regression is enabled.

Given an inference threshold \(\gamma\), the predicted close decision is
\[
\hat{a}_t=\mathbb{I}[\hat{A}_t\ge \gamma].
\]
Let \(a_t\in\{0,1\}\) denote the offline boundary monitoring label. Decision
accuracy is
\[
\mathrm{DecAcc}
=
\frac{1}{|\mathcal{T}|}
\sum_{t\in\mathcal{T}}
\mathbb{I}[\hat{a}_t=a_t].
\]

Close precision, recall, and F1 are
\[
\mathrm{Prec}_{\mathrm{close}}
=
\frac{\sum_t \mathbb{I}[\hat{a}_t=1\wedge a_t=1]}
{\sum_t \mathbb{I}[\hat{a}_t=1]+\epsilon},
\]
\[
\mathrm{Rec}_{\mathrm{close}}
=
\frac{\sum_t \mathbb{I}[\hat{a}_t=1\wedge a_t=1]}
{\sum_t \mathbb{I}[a_t=1]+\epsilon},
\]
\[
\mathrm{F1}_{\mathrm{close}}
=
\frac{2\mathrm{Prec}_{\mathrm{close}}\mathrm{Rec}_{\mathrm{close}}}
{\mathrm{Prec}_{\mathrm{close}}+\mathrm{Rec}_{\mathrm{close}}+\epsilon}.
\]

For the auxiliary peak head, the binary target is
\[
b_t=\mathbb{I}[A_t\ge \eta_{\mathrm{aux}}],
\]
and the predicted label is
\[
\hat{b}_t=\mathbb{I}[\hat{p}_t\ge 0.5].
\]
The auxiliary accuracy is
\[
\mathrm{AuxAcc}
=
\frac{1}{|\mathcal{T}|}
\sum_{t\in\mathcal{T}}
\mathbb{I}[\hat{b}_t=b_t].
\]

\paragraph{Player intention forecasting.}
For each evaluation point \(j\), let \(y_j\) be the target intent label,
\(\hat{y}_j\) be the predicted top-1 label, \(G_j^{(3)}\) be the GT top-3
intent set, and \(\hat{G}_j^{(3)}\) be the predicted top-3 intent set. Top-1
accuracy is
\[
\mathrm{Top1}
=
\frac{1}{M}
\sum_{j=1}^{M}
\mathbb{I}[\hat{y}_j=y_j].
\]
Soft accuracy is
\[
\mathrm{SoftAcc}
=
\frac{1}{M}
\sum_{j=1}^{M}
\mathbb{I}
\left[
\hat{y}_j\in G_j^{(3)}
\;\vee\;
y_j\in \hat{G}_j^{(3)}
\right].
\]

\paragraph{Advice timing retrieval.}
Let the ground-truth advice opportunities be
\(\mathcal{G}=\{g_j\}_{j=1}^{J}\), where each opportunity has a time span
\([s_j,e_j]\), center \(c_j\), and weight \(w_j\). Let predictions be
\(\mathcal{P}=\{p_i\}_{i=1}^{K}\), where each prediction has a timestamp
\(\hat{t}_i\) and span \([\hat{s}_i,\hat{e}_i]\).

The temporal IoU is
\[
\mathrm{IoU}_{ij}
=
\frac{
|[\hat{s}_i,\hat{e}_i]\cap[s_j,e_j]|
}{
|[\hat{s}_i,\hat{e}_i]\cup[s_j,e_j]|
}.
\]
The span distance is
\[
d^{\mathrm{span}}_{ij}
=
\begin{cases}
0, & [\hat{s}_i,\hat{e}_i]\cap[s_j,e_j]\neq\emptyset,\\
\min(|\hat{e}_i-s_j|,|\hat{s}_i-e_j|), & \text{otherwise}.
\end{cases}
\]
The span soft score is
\[
u^{\mathrm{span}}_{ij}
=
\exp\left(-d^{\mathrm{span}}_{ij}/\tau_{\mathrm{span}}\right).
\]

The center timing score is asymmetric for early and late predictions:
\[
u^{\mathrm{center}}_{ij}
=
\begin{cases}
1, & s_j\le \hat{t}_i\le e_j,\\
\exp(-\lambda_{\mathrm{early}}(s_j-\hat{t}_i)), & \hat{t}_i<s_j,\\
\exp(-\lambda_{\mathrm{late}}(\hat{t}_i-e_j)), & \hat{t}_i>e_j.
\end{cases}
\]

The temporal utility used in the implementation is
\[
u_{ij}
=
0.55\max(\mathrm{IoU}_{ij},u^{\mathrm{span}}_{ij})
+
0.45u^{\mathrm{center}}_{ij}.
\]

Predictions are matched to GT opportunities by maximum total utility under a
one-to-one matching constraint. Let \(\mathcal{M}\) denote the matched pairs
whose utility exceeds the score threshold. Weighted coverage at top \(K\) is
\[
\mathrm{WCov@K}
=
\frac{
\sum_{(j,i)\in\mathcal{M}} w_j
}{
\sum_{j=1}^{J}w_j
}.
\]

Temporal utility sum is
\[
\mathrm{TU\_sum}
=
\sum_{(j,i)\in\mathcal{M}}u_{ij}.
\]

Late-only latency is
\[
\mathrm{Latency}
=
\frac{1}{|\mathcal{M}|}
\sum_{(j,i)\in\mathcal{M}}
\max(0,\hat{t}_i-c_j).
\]

For ranking quality, each ranked prediction receives relevance
\[
\rho_i=\max_{j} w_j u_{ij}.
\]
The discounted cumulative gain is
\[
\mathrm{DCG@K}
=
\sum_{i=1}^{K}
\frac{\rho_i}{\log_2(i+1)}.
\]
Let \(\rho_i^\star\) be the same relevance scores sorted in descending order.
Then
\[
\mathrm{nDCG@K}
=
\frac{
\sum_{i=1}^{K}\frac{\rho_i}{\log_2(i+1)}
}{
\sum_{i=1}^{K}\frac{\rho_i^\star}{\log_2(i+1)}
}.
\]

\section{Hyperparameter Calibration and Sensitivity Analysis}
\label{app:sensitivity}

We examine sensitivity to the utility weights and the commitment threshold
\(\gamma\). Hyperparameters were calibrated using a small subset of the data
and downstream ground-truth signals. The deployed operating point is
\(\gamma=-0.47\), shared by the two downstream tasks.

\paragraph{Utility weights.}
\label{app:weight_sensitivity}
Table~\ref{tab:app_weight_sensitivity} summarizes rank agreement
under 12 utility reweighting configurations. For each diagnostic metric, the
top-ranked method remains unchanged in all 12 configurations. Worst-case
Kendall rank agreement ranges from 0.689 to 0.867, indicating that some
relative method orderings do change. These results support robustness of the
top-ranked method within the tested configurations, rather than invariance
of the full ranking or robustness to arbitrary utility definitions.

\begin{table}[tbp]
  \centering
  \small
  \caption{Method-ranking sensitivity across 12 utility reweightings.
  Agreement is measured relative to the nominal-weight ranking.
  The final column counts configurations that change the top-ranked method.}
  \label{tab:app_weight_sensitivity}
  \setlength{\tabcolsep}{4pt}
  \begin{tabular}{lcc}
    \toprule
    Ranking metric & Worst-case Kendall \(\tau\) & Top-method changes \\
    \midrule
    wSemC       & 0.867 & 0/12 \\
    wSep        & 0.733 & 0/12 \\
    wMBU        & 0.867 & 0/12 \\
    MBU\(^{+}\) & 0.689 & 0/12 \\
    \bottomrule
  \end{tabular}
\end{table}

\paragraph{Commitment threshold.}
\label{app:threshold_sensitivity}
Table~\ref{tab:app_threshold_sensitivity} shows that the threshold controls
episode granularity and the balance between coherence and closure utility.
Increasing \(\gamma\) from \(-0.60\) to \(-0.47\) reduces the reported episode
count from 176.2 to 15.5, while wMBU increases from 0.157 to 0.266 and the
positive-closure rate increases from 92.4\% to 99.2\%. Over the same sweep,
wSemC decreases from 0.542 to 0.401; wSep remains between 0.903 and 0.926
and is not monotonic. Thus, a coarser operating point improves the measured
closure utility at a cost in measured intra-episode coherence.
The deployed row matches the online-policy result in
Table~\ref{tab:online_oracle_gap}. Its wMBU is 0.021 below the offline
reference, with a positive-closure rate 0.2 percentage points lower.
This sweep characterizes the diagnostic trade-off; it does not establish
that every threshold is optimal for both downstream tasks.

\begin{table}[tbp]
  \centering
  \small
  \caption{Commitment-threshold sweep. Metric definitions and episode-count
  aggregation follow Table~\ref{tab:online_oracle_gap}. Bold marks the
  deployed operating point, not the best value in every column.
  The offline reference is included for comparison and is not a threshold setting.}
  \label{tab:app_threshold_sensitivity}
  \begin{tabular}{lrrrrr}
    \toprule
    \(\gamma\) & \#Ep & wSemC & wSep & wMBU & MBU\(^{+}\) (\%) \\
    \midrule
    \(-0.60\) & 176.2 & 0.542 & 0.903 & 0.157 & 92.4 \\
    \(-0.55\) &  84.2 & 0.505 & 0.925 & 0.198 & 94.6 \\
    \(-0.52\) &  45.7 & 0.449 & 0.926 & 0.228 & 98.1 \\
    \(-0.50\) &  28.7 & 0.421 & 0.922 & 0.241 & 98.7 \\
    \(\mathbf{-0.47}\) & \textbf{15.5} & \textbf{0.401} & \textbf{0.918}
      & \textbf{0.266} & \textbf{99.2} \\
    \midrule
    Offline reference & 35.3 & 0.532 & 0.931 & 0.287 & 99.4 \\
    \bottomrule
  \end{tabular}
\end{table}

\section{Utility Validation across Downstream Tasks}
\label{app:utility_validation}

We supplement the episode-quality diagnostics with correlations between
utility components and downstream performance. Both analyses use the
deployed setting \(\gamma=-0.47\), corresponding to the downstream results
in Table~\ref{tab:downstream_tasks}. The coefficients describe associations
with downstream outcomes and do not by themselves establish causal effects
of individual utility components.

\paragraph{Advice-timing retrieval.}
Table~\ref{tab:app_advice_correlation} relates utility components to three
retrieval outcomes. Positive-closure rate has the
largest positive coefficient for all three outcomes (0.57--0.62), followed
by coherence (0.44--0.50) and closure utility (0.23--0.26).
Separability has coefficients close to zero (\(-0.01\) to 0.06), and its
reported uncertainty intervals span zero for all three outcomes.
Accordingly, these results support the relevance of coherence and closure
quality to advice retrieval, but do not support a reliable positive
association between separability and advice-retrieval performance.

\begin{table}[tbp]
  \centering
  \small
  \caption{Reported correlations between utility components and advice-timing
  retrieval at \(\gamma=-0.47\). Point estimates are shown.}
  \label{tab:app_advice_correlation}
  \setlength{\tabcolsep}{3pt}
  \begin{tabular}{lrrr}
    \toprule
    Utility component & WCov@K & \(\mathrm{TU}_{\mathrm{sum}}\) & nDCG@K \\
    \midrule
    Coherence (wSemC)                 & 0.48 & 0.44 &  0.50 \\
    Separability (wSep)               & 0.06 & 0.03 & -0.01 \\
    Closure utility (wMBU)            & 0.26 & 0.23 &  0.24 \\
    Positive-closure rate (MBU\(^{+}\)) & 0.62 & 0.57 &  0.60 \\
    \bottomrule
  \end{tabular}
\end{table}

\paragraph{Intention forecasting.}
Table~\ref{tab:app_intent_correlation} shows a different pattern.
Separability correlates with Top-1 accuracy (0.432) and Soft accuracy
(0.430), while closure utility correlates with both outcomes (0.374 and
0.439). Coherence and positive-closure rate also have positive coefficients,
but their associations with Top-1 accuracy are smaller (0.132 and 0.160).
The utility components therefore have task-dependent associations:
separability is associated with intention forecasting, whereas coherence
and positive-closure rate are more strongly associated with advice retrieval
in the reported analyses. Together with the downstream results, this
supports evaluating episode formation through complementary utility
components and task outcomes, rather than treating a single diagnostic
metric as a universal measure of downstream effectiveness.

\begin{table}[tbp]
  \centering
  \small
  \caption{Reported correlations between utility components and intention
  forecasting at \(\gamma=-0.47\). Point estimates are shown.}
  \label{tab:app_intent_correlation}
  \begin{tabular}{lrr}
    \toprule
    Utility component & Top-1 accuracy & Soft accuracy \\
    \midrule
    Coherence (wSemC)                 & 0.132 & 0.234 \\
    Separability (wSep)               & 0.432 & 0.430 \\
    Closure utility (wMBU)            & 0.374 & 0.439 \\
    Positive-closure rate (MBU\(^{+}\)) & 0.160 & 0.343 \\
    \bottomrule
  \end{tabular}
\end{table}

\section{Statistical Analysis of Downstream Results}
\label{app:statistics}

We report the available significance comparisons for the downstream results
in Table~\ref{tab:downstream_tasks}. All p-values below are Holm-adjusted.
We distinguish improvements in point estimates from comparisons meeting
the \(p<0.05\) threshold after correction.

\paragraph{Intention forecasting.}
EpiStream obtains a Top-1 accuracy of 0.38.
Table~\ref{tab:app_intent_statistics} reports adjusted p-values against
four baselines. The improvements over DisPider, No-Memory, and Fixed-30s
meet the corrected significance threshold. The improvement over EM-Garde
does not (\(p=0.0625\)). These comparisons do not establish significance
against every baseline: in particular, no corresponding test result is
available here for OnGEBD, which obtains 0.32 Top-1 accuracy in the main
comparison. We therefore restrict the significance claim to the tested
comparisons that meet the corrected threshold.

\begin{table}[tbp]
  \centering
  \small
  \caption{Intention-forecasting Top-1 accuracy comparisons with EpiStream
  (0.38). DisPider is reported as 0.103 in the supplementary analysis and
  rounded to 0.10 in Table~\ref{tab:downstream_tasks}. Differences are
  approximate and rounded to two decimals, following the reported results.}
  \label{tab:app_intent_statistics}
  \begin{tabular}{lrrr}
    \toprule
    Baseline & Baseline Top-1 & \(\Delta\) Top-1 & Holm-adjusted \(p\) \\
    \midrule
    DisPider   & 0.103 & +0.28 & 0.0066 \\
    No-Memory  & 0.08  & +0.30 & 0.0026 \\
    Fixed-30s  & 0.22  & +0.16 & 0.0078 \\
    EM-Garde   & 0.28  & +0.10 & 0.0625 \\
    \bottomrule
  \end{tabular}
\end{table}

\paragraph{Advice-timing retrieval.}
Table~\ref{tab:app_advice_statistics} compares EpiStream with DisPider,
the strongest competing method on the three reported retrieval-quality
metrics. The WCov@K improvement is 0.041 and the nDCG@K improvement is
0.015; both have reported Holm-adjusted \(p=4\times10^{-5}\).
The \(\mathrm{TU}_{\mathrm{sum}}\) point estimate increases by 0.162,
but the adjusted \(p=0.080\) does not meet the 0.05 threshold.
Thus, the available tests support improved coverage and ranking quality
against DisPider, while the total-utility gain remains descriptive.
The lower latency in the main table (14.4\,s versus 17.7\,s) is also
reported descriptively because no corresponding significance test is
included in this analysis.

\begin{table}[tbp]
  \centering
  \small
  \caption{Advice-timing retrieval comparisons with DisPider. Point estimates
  match Table~\ref{tab:downstream_tasks}; \(\Delta\) is EpiStream minus
  DisPider, computed from the displayed scores.}
  \label{tab:app_advice_statistics}
  \begin{tabular}{lrrrr}
    \toprule
    Metric & EpiStream & DisPider & \(\Delta\) & Holm-adjusted \(p\) \\
    \midrule
    WCov@K                       & 0.466 & 0.425 & +0.041 & \(4\times10^{-5}\) \\
    nDCG@K                       & 0.509 & 0.494 & +0.015 & \(4\times10^{-5}\) \\
    \(\mathrm{TU}_{\mathrm{sum}}\) & 1.855 & 1.693 & +0.162 & 0.080 \\
    \bottomrule
  \end{tabular}
\end{table}

\section{Discussion and Limitations}
\paragraph{Impact.}
More broadly, we argue that dense-action streaming understanding represents an important yet underexplored frontier for video foundation models. Real-world streaming environments—including gameplay, sports broadcasts, embodied agents, and human activity streams—are dominated by continuous, overlapping dynamics rather than clean event boundaries. However, existing video understanding benchmarks and methods remain largely centered on offline recognition, boundary annotation, or short-horizon question answering, leaving the problem of utility-aware episodic abstraction insufficiently studied. In particular, the community currently lacks (i) standardized datasets for long-form dense-action streaming understanding, (ii) unified metrics for evaluating semantic episode quality, and (iii) principled formulations for deciding when streaming observations should be committed into memory for downstream reasoning.

We view EpiStream not merely as a method, but as a general paradigm for utility-aware episodic abstraction in streaming video understanding. By reframing temporal abstraction as a causal commit decision problem, our framework provides a transferable formulation that can extend beyond gameplay to broader dense-action domains such as sports analytics, embodied AI, and interactive assistants. We hope this work motivates future research on utility-aware streaming memory construction, semantic episode evaluation, and long-horizon reasoning in continuous video streams.

\paragraph{Limitations.}

Although EpiStream provides a general utility-aware formulation for dense-action stream understanding, several limitations remain. First, the relative weighting of semantic coherence, inter-episode distinctiveness, and task-specific utility is inherently task-oriented and may need to be adjusted for different downstream objectives. For example, intention forecasting may favor episodes with stable latent states, while proactive advice retrieval may emphasize timely closure and decision-critical evidence. Second, applying EpiStream to new DAS scenarios may require a small amount of task-oriented adaptation data to calibrate the utility function and distill the corresponding online commitment policy, since useful episode boundaries can vary across domains and downstream reasoning goals. Finally, high-quality dense-action streaming datasets with reliable temporal signals and downstream utility annotations are still limited. Future work will include more diverse DAS data sources and task settings to further evaluate and extend utility-aware episode formation.


\newpage
\section*{NeurIPS Paper Checklist}

\begin{enumerate}

\item {\bf Claims}
    \item[] Question: Do the main claims made in the abstract and introduction accurately reflect the paper's contributions and scope?
    \item[] Answer: \answerYes{} 
    \item[] Justification: The abstract and introduction state that EpiStream addresses utility-aware semantic episode formation in dense-action streams, rather than generic event boundary detection. The claims are supported by the methodology, which formalizes causal episode commitment and utility-derived advantage learning, and by experiments showing improvements in semantic coherence and downstream tasks under fixed video and memory budgets.

\item {\bf Limitations}
    \item[] Question: Does the paper discuss the limitations of the work performed by the authors?
    \item[] Answer: \answerYes{} 
    \item[] Justification: The paper discusses the current scope of the evaluation as long-form FPS gameplay streams with synchronized telemetry and notes that future work will extend the formulation to broader dense-action domains such as driving, sports commentary, and human activity streams. The method also explicitly assumes access to privileged telemetry for offline utility construction, while the online policy itself uses only causal video observations at inference time.

\item {\bf Theory assumptions and proofs}
    \item[] Question: For each theoretical result, does the paper provide the full set of assumptions and a complete (and correct) proof?
    \item[] Answer: \answerNA{} 
    \item[] Justification: The paper does not present formal theoretical theorems or proofs. The mathematical derivations are used to define the utility function, marginal boundary utility, close-versus-continue advantage, and the probabilistic interpretation of the joint advantage-peak objective; these definitions and assumptions are provided in the methodology and Appendix ~\ref{app:offline_utility_construction}.

    \item {\bf Experimental result reproducibility}
    \item[] Question: Does the paper fully disclose all the information needed to reproduce the main experimental results of the paper to the extent that it affects the main claims and/or conclusions of the paper (regardless of whether the code and data are provided or not)?
    \item[] Answer: \answerYes{} 
    \item[] Justification: The paper specifies the dataset setting, telemetry-derived supervision, causal inference constraint, baseline groups, semantic episode metrics, downstream evaluation metrics, and the fixed-budget comparison protocol. The method section provides the online episode formation algorithm, utility construction, advantage target construction, and training objective needed to reproduce the core experimental logic.

\item {\bf Open access to data and code}
    \item[] Question: Does the paper provide open access to the data and code, with sufficient instructions to faithfully reproduce the main experimental results, as described in supplemental material?
    \item[] Answer: \answerYes{} 
    \item[] Justification: The current submission does not provide open access to the gameplay streams, synchronized telemetry, or training/evaluation code. This is primarily because the experiments rely on long-form gameplay videos paired with synchronized telemetry logs; the paper instead provides detailed descriptions of the utility construction, online policy, evaluation metrics, and fixed-budget protocol to support reproducibility.

\item {\bf Experimental setting/details}
    \item[] Question: Does the paper specify all the training and test details (e.g., data splits, hyperparameters, how they were chosen, type of optimizer) necessary to understand the results?
    \item[] Answer: \answerYes{} 
    \item[] Justification: The paper describes the long-form FPS gameplay setting, telemetry signals, causal inference constraint, baselines, metrics, downstream tasks, and fixed memory-budget protocol. It also provides the training objective for joint advantage regression and peak prediction, including the advantage target, peak target, loss terms, and inference thresholding rule.

\item {\bf Experiment statistical significance}
    \item[] Question: Does the paper report error bars suitably and correctly defined or other appropriate information about the statistical significance of the experiments?
    \item[] Answer: \answerNo{} 
    \item[] Justification: The paper reports aggregate results across 60 dense-action sessions but does not currently include error bars, confidence intervals, or statistical significance tests. The reported results should therefore be interpreted as aggregate empirical comparisons rather than formal significance-tested estimates.

\item {\bf Experiments compute resources}
    \item[] Question: For each experiment, does the paper provide sufficient information on the computer resources (type of compute workers, memory, time of execution) needed to reproduce the experiments?
    \item[] Answer: \answerYes{} 
    \item[] Justification: The paper specifies the exact compute resources, GPU type, memory, training time, or total compute used for policy learning and downstream evaluation. These details are added in the appendix or supplemental material for improved reproducibility.
    
\item {\bf Code of ethics}
    \item[] Question: Does the research conducted in the paper conform, in every respect, with the NeurIPS Code of Ethics \url{https://neurips.cc/public/EthicsGuidelines}?
    \item[] Answer: \answerYes{} 
    \item[] Justification: The work studies utility-aware temporal abstraction for video streams and does not involve human-subject experiments, personally identifiable information, or deployment on real users. The evaluation uses gameplay videos and telemetry-derived signals for offline supervision, and the online policy does not access privileged telemetry at inference time.

\item {\bf Broader impacts}
    \item[] Question: Does the paper discuss both potential positive societal impacts and negative societal impacts of the work performed?
    \item[] Answer: \answerNo{} 
    \item[] Justification: The current paper focuses primarily on the technical formulation and empirical evaluation and does not include a dedicated broader impact discussion. Potential positive impacts include improved streaming video understanding and real-time decision support, while potential risks include misuse for surveillance-like monitoring or over-reliance on automated advice in high-stakes streaming environments; these should be discussed in a broader impact or limitations paragraph.
    
\item {\bf Safeguards}
    \item[] Question: Does the paper describe safeguards that have been put in place for responsible release of data or models that have a high risk for misuse (e.g., pre-trained language models, image generators, or scraped datasets)?
    \item[] Answer: \answerNA{} 
    \item[] Justification: The paper does not release a high-risk pretrained generative model, scraped internet dataset, or user-facing deployment system. If gameplay streams, telemetry logs, or trained checkpoints are released later, the authors should document usage restrictions, anonymization procedures, and any domain-specific safeguards.

\item {\bf Licenses for existing assets}
    \item[] Question: Are the creators or original owners of assets (e.g., code, data, models), used in the paper, properly credited and are the license and terms of use explicitly mentioned and properly respected?
    \item[] Answer: \answerNo{} 
    \item[] Justification: The paper cites prior methods and benchmarks, but the current version does not explicitly state the licenses or terms of use for existing assets such as the VLM backbone, gameplay video sources, telemetry extraction tools, or baseline implementations. These license details should be added to the appendix or supplementary material.

\item {\bf New assets}
    \item[] Question: Are new assets introduced in the paper well documented and is the documentation provided alongside the assets?
    \item[] Answer: \answerYes{} 
    \item[] Justification: The paper introduces the DAS-Utility evaluation protocol and describes the long-form gameplay streams, synchronized telemetry signals, utility-derived teaching targets, semantic episode metrics, and downstream tasks. If the benchmark assets are released, the documentation should additionally include data format, preprocessing scripts, telemetry schema, license, and intended-use statements.

\item {\bf Crowdsourcing and research with human subjects}
    \item[] Question: For crowdsourcing experiments and research with human subjects, does the paper include the full text of instructions given to participants and screenshots, if applicable, as well as details about compensation (if any)? 
    \item[] Answer: \answerNA{} 
    \item[] Justification: The paper does not conduct research with human subjects, collect participant responses, or run user studies. Therefore IRB approval or equivalent human-subjects review is not applicable.

\item {\bf Institutional review board (IRB) approvals or equivalent for research with human subjects}
    \item[] Question: Does the paper describe potential risks incurred by study participants, whether such risks were disclosed to the subjects, and whether Institutional Review Board (IRB) approvals (or an equivalent approval/review based on the requirements of your country or institution) were obtained?
    \item[] Answer: \answerNA{} 
    \item[] Justification: This work does not involve crowdsourcing, user studies, or experiments with human subjects. The dataset consists of gameplay videos and synchronized telemetry/event logs used for offline utility construction and evaluation, without collecting personally identifiable information or interacting with human participants

\item {\bf Declaration of LLM usage}
    \item[] Question: Does the paper describe the usage of LLMs if it is an important, original, or non-standard component of the core methods in this research? Note that if the LLM is used only for writing, editing, or formatting purposes and does \emph{not} impact the core methodology, scientific rigor, or originality of the research, declaration is not required.
    \item[] Answer: \answerYes{} 
    \item[] Justification: The method uses a VLM backbone with a lightweight decision head for causal commitment policy learning, and the experiments use a downstream LLM/VLM under a fixed memory and video budget. Because LLM/VLM components are central to the policy and downstream evaluation, the paper should clearly specify the model family, prompting/state construction, and whether any LLM was used for data construction or evaluation.

\end{enumerate}

\end{document}